%% file: neurips_2026.tex
\documentclass{article}

\usepackage[preprint]{neurips_2026}

\usepackage{hyperref}
\usepackage{url}

\usepackage{amsmath}
\usepackage[accsupp]{axessibility}
\usepackage{lineno}
\usepackage{etoolbox}

\usepackage{wrapfig}
\usepackage{graphicx}
\usepackage{grffile}

\usepackage[page, header, toc, page]{appendix} 
\usepackage{hhline}
\usepackage{caption}
\usepackage{subcaption}
\usepackage{enumitem}
\usepackage[utf8]{inputenc} 
\usepackage[T1]{fontenc}    
\usepackage{hyperref}       
\usepackage{url}            
\usepackage{booktabs}       
\usepackage{amsfonts}       
\usepackage{nicefrac}       
\usepackage{microtype}      
\usepackage{xcolor}         
\usepackage{multirow}
\usepackage{amssymb}
\usepackage{amsmath}
\usepackage{bbding}
\usepackage{makecell}
\usepackage{colortbl}  
\usepackage{array}  
\usepackage{arydshln}
\usepackage{subcaption,siunitx,booktabs}
\usepackage{empheq, nccmath}
\usepackage{tabularx}
\usepackage{mathtools}
\usepackage{adjustbox}
\usepackage{graphicx}
\usepackage{grffile}
\usepackage{float}
\usepackage{minitoc}

\usepackage{algorithm}
\usepackage{algorithmic}
\usepackage[capitalize,noabbrev]{cleveref}
\usepackage{verbatim}
\usepackage{minted}
\usepackage{cuted}
\usepackage{capt-of}
\usepackage{tabularx}
\usepackage{tabu}
\usepackage{tikz,pgfplots}
\usepackage{dsfont}
\usepackage{etoc}
\usepackage{titletoc}
\usetikzlibrary{arrows.meta, positioning, calc, shadows}

\definecolor{arrowgray}{RGB}{150,150,150}
\definecolor{profgreen}{RGB}{200,30,30}
\definecolor{ForestGreen}{RGB}{34,139,34}

\setminted[python]{breaklines}
\newcommand{\myparagraph}[1]{\vspace{2pt}\noindent{\bf #1}}

\usepackage{pifont}

\newcommand{\cmark}{\ding{51}}
\newcommand{\xmark}{\ding{55}}

\definecolor{checkgreen}{RGB}{0,130,80}
\definecolor{crossred}{RGB}{190,55,55}

\newcommand{\gcmark}{\textcolor{checkgreen}{\cmark}}
\newcommand{\rxmark}{\textcolor{crossred}{\xmark}}

\title{Training-Free Global Geometric Association for 4D LiDAR Panoptic Segmentation}

\author{
  {\bfseries
  Gyeongrok Oh\textsuperscript{\normalfont 1} \hspace{0.5em}
  Youngdong Jang\textsuperscript{\normalfont 1} \hspace{0.5em}
  Jonghyun Choi\textsuperscript{\normalfont 2} \hspace{0.5em}
  Suk-Ju Kang\textsuperscript{\normalfont 3} \hspace{0.5em}
  } \\
  {\bfseries
  Guang Lin\textsuperscript{\normalfont 4} \hspace{0.5em}
  Sangpil Kim\textsuperscript{\normalfont $\dagger$,1}
  } \\
  \\
  \textsuperscript{\normalfont 1} Korea University \quad
  \textsuperscript{\normalfont 2} Hyundai Motor Company \quad
  \textsuperscript{\normalfont 3} Sogang University \quad 
  \textsuperscript{\normalfont 4} Purdue University
}

\input{preamble}
\begin{document}

\maketitle
\begingroup
\renewcommand{\thefootnote}{}
\renewcommand{\footnoterule}{}
\footnotetext{$\dagger$ Corresponding author.}
\endgroup

\input{sec/0_abstract}    
\input{sec/1_intro}
\input{sec/2_related}

\input{sec/3_method}

\input{sec/4_experiment}

\input{sec/5_conclusion}

\bibliographystyle{unsrt}
\bibliography{neurips_2026}

\appendix
\input{sec/X_suppl}

\end{document}

%% file: preamble.tex
\newcommand{\authcount}[1]{}

\usepackage{etoc}
\usepackage{diagbox} 
\definecolor{others}{RGB}{0, 0, 0}
\definecolor{nbarrier}{RGB}{255, 120, 50}
\definecolor{nbicycle}{RGB}{255, 192, 203}
\definecolor{nbus}{RGB}{255, 255, 0}
\definecolor{ncar}{RGB}{0, 150, 245}
\definecolor{nconstruct}{RGB}{0, 255, 255}
\definecolor{nmotor}{RGB}{200, 180, 0}
\definecolor{npedestrian}{RGB}{255, 0, 255}
\definecolor{ntraffic}{RGB}{255, 240, 150}
\definecolor{ntrailer}{RGB}{135, 60, 0}
\definecolor{ntruck}{RGB}{255, 0, 0}
\definecolor{ndriveable}{RGB}{213, 213, 213}
\definecolor{nother}{RGB}{139, 137, 137}
\definecolor{nsidewalk}{RGB}{75, 0, 75}
\definecolor{nterrain}{RGB}{150, 240, 80}
\definecolor{nmanmade}{RGB}{160, 32, 240}
\definecolor{nvegetation}{RGB}{0, 175, 0}

\definecolor{car}{rgb}{0.39215686, 0.58823529, 0.96078431}
\definecolor{bicycle}{rgb}{0.39215686, 0.90196078, 0.96078431}
\definecolor{motorcycle}{rgb}{0.11764706, 0.23529412, 0.58823529}
\definecolor{truck}{rgb}{0.31372549, 0.11764706, 0.70588235}
\definecolor{other-vehicle}{rgb}{0.39215686, 0.31372549, 0.98039216}
\definecolor{person}{rgb}{1.        , 0.11764706, 0.11764706}
\definecolor{bicyclist}{rgb}{1.        , 0.15686275, 0.78431373}
\definecolor{motorcyclist}{rgb}{0.58823529, 0.11764706, 0.35294118}
\definecolor{road}{rgb}{1.        , 0.        , 1.        }
\definecolor{parking}{rgb}{1.        , 0.58823529, 1.        }
\definecolor{sidewalk}{rgb}{0.29411765, 0.        , 0.29411765}
\definecolor{other-ground}{rgb}{0.68627451, 0.        , 0.29411765}
\definecolor{building}{rgb}{1.        , 0.78431373, 0.        }
\definecolor{fence}{rgb}{1.        , 0.47058824, 0.19607843}
\definecolor{vegetation}{rgb}{0.        , 0.68627451, 0.        }
\definecolor{trunk}{rgb}{0.52941176, 0.23529412, 0.        }
\definecolor{terrain}{rgb}{0.58823529, 0.94117647, 0.31372549}
\definecolor{pole}{rgb}{1.        , 0.94117647, 0.58823529}
\definecolor{traffic-sign}{rgb}{1.        , 0.        , 0.    }   

\definecolor{s_car}{rgb}{0.39215686, 0.58823529, 0.96078431}
\definecolor{s_bicycle}{rgb}{0.39215686, 0.90196078, 0.96078431}
\definecolor{s_motorcycle}{rgb}{0.11764706, 0.23529412, 0.58823529}
\definecolor{s_truck}{rgb}{0.31372549, 0.11764706, 0.70588235}
\definecolor{s_othervehicle}{rgb}{0.39215686, 0.31372549, 0.98039216}
\definecolor{s_person}{rgb}{1.        , 0.11764706, 0.11764706}
\definecolor{s_bicyclist}{rgb}{1.        , 0.15686275, 0.78431373}
\definecolor{s_motorcyclist}{rgb}{0.58823529, 0.11764706, 0.35294118}
\definecolor{s_road}{rgb}{1.        , 0.        , 1.        }
\definecolor{s_parking}{rgb}{1.        , 0.58823529, 1.        }
\definecolor{s_sidewalk}{rgb}{0.29411765, 0.        , 0.29411765}
\definecolor{s_otherground}{rgb}{0.68627451, 0.        , 0.29411765}
\definecolor{s_building}{rgb}{1.        , 0.78431373, 0.        }
\definecolor{s_fence}{rgb}{1.        , 0.47058824, 0.19607843}
\definecolor{s_vegetation}{rgb}{0.        , 0.68627451, 0.        }
\definecolor{s_trunk}{rgb}{0.52941176, 0.23529412, 0.        }
\definecolor{s_terrain}{rgb}{0.58823529, 0.94117647, 0.31372549}
\definecolor{s_pole}{rgb}{1.        , 0.94117647, 0.58823529}
\definecolor{s_trafficsign}{rgb}{1.        , 0.        , 0.        }
\definecolor{s_otherstructure}{rgb}{0.98039215, 0.58823529, 0.}
\definecolor{s_otherobject}{rgb}{0.19607843, 1.        , 1.        }

\makeatletter
\newcommand{\barrier@nuscenesfreq}{11.79}
\newcommand{\bicycle@nuscenesfreq}{0.18}
\newcommand{\bus@nuscenesfreq}{5.83}
\newcommand{\car@nuscenesfreq}{48.27}
\newcommand{\construction@nuscenesfreq}{1.92}
\newcommand{\motorcycle@nuscenesfreq}{0.54}
\newcommand{\pedestrian@nuscenesfreq}{2.93}
\newcommand{\trafficcone@nuscenesfreq}{0.93}
\newcommand{\trailer@nuscenesfreq}{6.22}
\newcommand{\truck@nuscenesfreq}{20.07}
\newcommand{\driveable@nuscenesfreq}{28.64}
\newcommand{\other@nuscenesfreq}{0.77}
\newcommand{\sidewalk@nuscenesfreq}{6.34}
\newcommand{\terrain@nuscenesfreq}{6.35}
\newcommand{\manmade@nuscenesfreq}{16.10}
\newcommand{\vegetation@nuscenesfreq}{11.08}
\newcommand{\nuscenesfreq}[1]{{\csname #1@nuscenesfreq\endcsname}}

\newcommand{\car@semkitfreq}{3.92}
\newcommand{\bicycle@semkitfreq}{0.03}
\newcommand{\motorcycle@semkitfreq}{0.03}
\newcommand{\truck@semkitfreq}{0.16}
\newcommand{\othervehicle@semkitfreq}{0.20}
\newcommand{\person@semkitfreq}{0.07}
\newcommand{\bicyclist@semkitfreq}{0.07}
\newcommand{\motorcyclist@semkitfreq}{0.05}
\newcommand{\road@semkitfreq}{15.30}  %
\newcommand{\parking@semkitfreq}{1.12}
\newcommand{\sidewalk@semkitfreq}{11.13}  %
\newcommand{\otherground@semkitfreq}{0.56}
\newcommand{\building@semkitfreq}{14.1}  %
\newcommand{\fence@semkitfreq}{3.90}
\newcommand{\vegetation@semkitfreq}{39.3}  %
\newcommand{\trunk@semkitfreq}{0.51}
\newcommand{\terrain@semkitfreq}{9.17} %
\newcommand{\pole@semkitfreq}{0.29}
\newcommand{\trafficsign@semkitfreq}{0.08}
\newcommand{\semkitfreq}[1]{{\csname #1@semkitfreq\endcsname}}

\makeatletter
\let\@makespecialcolbox\@make@specialcolbox
\makeatother

%% file: sec/0_abstract.tex
\begin{abstract}
Dominant paradigms for 4D LiDAR panoptic segmentation are usually required to train deep neural networks with large superimposed point clouds or design dedicated modules for instance association.
However, these approaches perform redundant point processing and consequently become computationally expensive, yet still overlook the rich geometric priors inherently provided by raw point clouds.
To this end, we introduce \textsc{Geo-4D}, a simple yet effective training-free framework that unifies spatial and temporal reasoning, enabling holistic LiDAR perception over long time horizons.
Specifically, we propose a global geometric association strategy that establishes consistent instance correspondences by estimating an optimal transformation between instance-level point sets.
To mitigate instability caused by structural inconsistencies in point cloud observations, we propose a global geometry-aware soft matching mechanism that enforces spatially coherent point-wise correspondences grounded in the spatial distribution of instance point sets.
Furthermore, our carefully designed pipeline, which considers three instance types—static, dynamic, and missing—offers computational efficiency and occlusion-aware matching.
Our extensive experiments across both SemanticKITTI and nuScenes demonstrate that our method consistently outperforms state-of-the-art approaches, even without additional training or extra point cloud inputs.
These results underscore the potential of inherent geometric priors in point clouds as an efficient solution for maintaining consistent instance identities over time.
\end{abstract}

%% file: sec/1_intro.tex
\section{Introduction}
\label{sec:intro}

Autonomous driving systems heavily depend on precise perception of dynamic 3D environments with geometrically accurate measurements captured by LiDAR sensors. 
In particular, LiDAR segmentation plays a crucial role in enabling comprehensive point-wise understanding.
In 3D environments, LiDAR panoptic segmentation~\citep{kirillov2019panoptic, milioto2020lidar} originates from semantic and instance segmentation, which jointly identify class categories and instance IDs. 
Along with the recent significant advances in frame-wise perception, 4D LiDAR panoptic segmentation~\citep{Aygun_2021_CVPR} has emerged to capture temporal relationships in a holistic manner.
In this context, ensuring temporal continuity is a principal purpose in 4D LiDAR panoptic segmentation, even under instance motions.

With its potential benefits, as shown in Table~\ref{tab:conceptual_comparison}, the mainstream of prior works can be categorized into three discrete paradigms: 1) IoU-based, 2) Query-propagated, and 3) Detect \& Track association.
One solution~\citep{Aygun_2021_CVPR, kreuzberg20224d, zhu20234d, yilmaz2024mask4former, 10380457}, superimposing point clouds, is widely adopted for establishing 4D spatio-temporal volume. It enables consistent instance tracking over time by measuring overlapping regions.
However, despite its simplicity, stacking point clouds (i.e., increasing the number of points per batch) leads to a quadratic growth in computational overhead, which is an inherent challenge in point cloud processing~\citep{liu2023flatformer, lai2022stratified,wu2022point,wu2024point}.

\begin{wraptable}{r}{0.48\columnwidth}
    \vspace{-8pt}
    \centering
    \caption{\small Conceptual comparison of 4D LiDAR panoptic segmentation paradigms.}
    \label{tab:conceptual_comparison}
    
    \vspace{-3pt}
    \scriptsize
    \setlength{\tabcolsep}{2.5pt}
    \renewcommand{\arraystretch}{1.05}

    \resizebox{\linewidth}{!}{
    \begin{tabular}{lccc}
        \toprule
        \textbf{Strategy}
        & \shortstack{\textbf{Frame-wise}\\\textbf{Perception}}
        & \shortstack{\textbf{No Temporal}\\\textbf{Training}}
        & \shortstack{\textbf{Off-the-shelf}\\\textbf{3D Perception Model}} \\
        \midrule

        IoU-based
        & \rxmark
        & \rxmark
        & \rxmark \\

        Query Prop.
        & \gcmark
        & \rxmark
        & \rxmark \\

        Detect \& Track
        & \gcmark
        & \rxmark
        & \gcmark \\

        \midrule
        \textbf{Geo-4D (Ours)}
        & \gcmark
        & \gcmark
        & \gcmark \\
        \bottomrule
    \end{tabular}
    }
    \vspace{-8pt}
\end{wraptable}

An alternative direction~\citep{marcuzzi2023mask4d,athar20234d, marcuzzi2022contrastive} seeks to mitigate this issue by treating each scan independently, as demonstrated in the 2\textsuperscript{nd} and 3\textsuperscript{rd} rows of
Table~\ref{tab:conceptual_comparison}.
They are primarily derived from the two major paradigms of multi-object tracking, tracking-by-detection~\citep{bewley2016simple, wojke2017simple, zhang2022bytetrack,qin2024towards} and query propagation~\citep{zeng2022motr, zhang2023motrv2, meinhardt2022trackformer}, to assign consistent IDs for instances.
Both approaches differ substantially in training strategies, yet commonly suffer from error accumulation~\citep{zhang2023motrv2, segu2024samba} and reliance on costly and dataset-dependent training.
Consequently, we argue that this suboptimal design fails to fully exploit the strong perceptual capabilities of 3D panoptic networks, which naturally raises the following question: \textit{Can instance association be achieved efficiently using only a 3D panoptic network?}

To answer this, we propose a global geometric association strategy for consistent and reliable instance correspondence, even without any training process.
Iterative Closest Point~(ICP) is widely used for point cloud registration~\citep{wang2019deep, zhang20233d, choy2020deep}, but its applicability beyond that has been largely underexplored.  
Here, we integrate ICP into the 4D LiDAR panoptic segmentation framework, thereby enabling explicit instance correspondence through geometric alignment under estimated inter-scan transformations.
Additionally, we introduce a novel paradigm for point-wise correspondence, Global Geometry-aware Soft Matching~(GGSM), enhancing the robustness to structural noise by enforcing instance-aware geometric correspondences.
Traditional ICP performs well on point cloud pairs under ideal conditions, but it relies on nearest-neighbor point correspondences.
In real-world noisy measurements acquired from on-vehicle sensors, such point-wise matching often yields incorrect transformations and ultimately leads to erroneous instance associations, as it ignores the global instance structure.
In contrast, our GGSM allows reliable correspondences based on the overall geometric context rather than local proximity.
Specifically, it identifies point pairs that minimize the transport cost by solving the earth mover’s problem~\citep{cuturi2013sinkhorn} between two instance point set distributions. 
As a result, instance association shifts from greedy local pairing to globally consistent alignment, yielding more stable and accurate matching even under challenging environments.

In addition, in real-world observations, instances may remain stationary, move across frames, or temporarily disappear due to occlusion or sensing limitations.
Motivated by these instance states, we classify instances into three types: static, dynamic, and missing.
For static instances, we apply statistics-based geometric filtering using the mean and covariance of their point distributions to preserve spatial consistency across frames.
This filtering step also significantly reduces the number of matching candidates before applying global geometric association to dynamic instances, thereby ensuring computational efficiency.
Moreover, we incorporate a short-term memory bank to recover temporarily unobservable instances. 
We retain only unmatched instances within a short-term window, preserving temporal consistency with minimal memory overhead.

To wrap it up, we named our overall pipeline as \textbf{\textsc{Geo-4D}} and demonstrate its effectiveness through extensive experiments on the SemanticKITTI~\citep{behley2019semantickitti} and nuScenes~\citep{fong2022panoptic} benchmarks. Note that \textsc{Geo-4D} achieves superior performance over existing state-of-the-art methods and ranks 1st on the SemanticKITTI test leaderboard, even under the minimal setting of a single scan and without any additional training. 
The contributions of this work are summarized as follows:
(1) we introduce \textsc{Geo-4D}, a training-free 4D LiDAR panoptic segmentation pipeline that can be flexibly integrated with various 3D panoptic segmentation networks; 
(2) we propose a global geometric association strategy with instance state conditioned solutions to improve association robustness while efficiently handling computational complexity and occlusion; and 
(3) we demonstrate the superiority of \textsc{Geo-4D} through extensive experiments and in-depth analyses on challenging benchmarks.

%% file: sec/2_related.tex
\section{Related Work}
\label{sec:related}

\myparagraph{LiDAR Panoptic Segmentation.}
LiDAR panoptic segmentation handles two different basis tasks, semantic and instance segmentation.
Proposal-based methods~\cite{liu2019end, mohan2021efficientps, xiong2019upsnet} first identify \textit{thing} points and then group them into instances, whereas proposal-free methods~\citep{hong2021lidar, razani2021gp, zhou2021panopticp, li2022panoptic} directly predict semantic labels and class agnostic instance segments.
4D LiDAR panoptic segmentation further requires instance identities to remain consistent across sequential scans.
Most existing methods~\citep{zhang2025zero, Aygun_2021_CVPR, kreuzberg20224d, zhu20234d, yilmaz2024mask4former, 10380457} aggregate multiple consecutive scans and associate instances based on overlapping regions. Alternatively, CA-Net~\citep{marcuzzi2022contrastive} uses feature similarity, while Mask4D~\citep{marcuzzi2023mask4d} maintains identities by iteratively updating track queries.
These approaches often incur additional computational cost or depend on specific network designs.
In contrast, we propose \textsc{Geo-4D} that is a flexible, efficient, and well-balanced instance association through direct geometric registration.

\myparagraph{Multi-Object Tracking~(MOT).}
Tracking-by-detection~(TBD)~\citep{bewley2016simple, wojke2017simple, zhang2022bytetrack,qin2024towards} has become the \textit{de facto} standard due to its simplicity and the rapid progress of detectors. Here, the motion model~(e.g., Kalman Filter~\citep{kalman1960new}) plays a key role by estimating future object states to maintain tracking continuity. 
Recent DETR-based methods~\citep{zeng2022motr, zhang2023motrv2, meinhardt2022trackformer} instead perform detection and tracking jointly with query-based representations, and similar paradigms have been extended to 3D tracking~\citep{weng2020ab3dmot, zhou2020tracking, pang2022simpletrack, zhang2022mutr3d}.
Unlike conventional MOT, however, 4D LiDAR panoptic segmentation requires point-level scene understanding rather than bounding box tracking.
\textsc{Geo-4D} follows the TBD paradigm while grounding instance association in the geometric structure of the underlying point clouds.

\myparagraph{Iterative Closest Point~(ICP).}
ICP~\citep{besl1992method} is a classical point set registration method with numerous extensions~\citep{bustos2017guaranteed, li2021practical, rusu2009fast, bouaziz2013sparse, rusinkiewicz2001efficient, segal2009generalized, yang2015go}.
It has also been incorporated into deep learning–based applications~\citep{wang2019deep, zhang20233d, choy2020deep}.
ICP-Flow~\citep{lin2024icp}, which estimates point-wise scene flow between adjacent scans using instance-level clustering, is closely related to our work.
However, they only consider point-wise motion estimation between two temporally adjacent scans by leveraging well-separated instance clusters.
In contrast, we use ICP for persistent instance association across sequential scans.
Remarkably, our Global Geometry-aware Soft Matching enables instance centric association, even when instance point sets are noisy, imprecise, or fragmented.

%% file: sec/3_method.tex
\section{Method}
\label{sec:method}

\subsection{Preliminary}
\label{sec:preliminary}

\myparagraph{Iterative Closest Point~(ICP).}
\label{sec:cip}
Conventionally, for point cloud registration, ICP estimates the optimal transformation by iteratively minimizing the difference between source and target point sets. Given two point sets $\mathcal{P}_\text{src} = \{p_1,...,p_I\}$ and $\mathcal{P}_\text{dst} = \{q_1,...,q_J\}$ in $\mathbb{R}^{3}$, it starts with the initial guess of rigid transformation ($\mathbf{R}^{(0)}, \mathbf{t}^{(0)}$), where the rotation $\mathbf{R}\in\text{SO}(3)$ and translation $\mathbf{t}\in\mathbb{R}^{3}$, which satisfies the following formula:
\begin{equation}
\label{eq-icp}
     q \approx \mathbf{R}p+\mathbf{t}\;.
\end{equation}
Here, in order to compute the alignment error, pairwise correspondences are established using a nearest-neighbor search algorithm (e.g., KD-tree~\citep{bentley1975multidimensional}) as:
\begin{equation}
\label{eq-comb_nn}
    \mathcal{C}^{(k)} = \{(p,\hat{q})\;|\; \hat{q} =  {\underset{q\in\mathcal{P}_\text{dst}}{\operatorname*{arg\,min}}||\mathbf{R}^{(k)}p + \mathbf{t}^{(k)} - q||,\;\forall p \in \mathcal{P}_\text{src}}\},
\end{equation}
where $k$ denotes the iteration.
Classical ICP adopts $l_2$ distance as the error metric between the matched point pairs. 
Ultimately, the optimal rigid transformation is obtained by iteratively solving a series of least-squares problems as: 
\begin{equation}
\mathbf{R}^{(k+1)}, \mathbf{t}^{(k+1)} = \underset{\mathbf{R}, \mathbf{t}}{\operatorname*{arg\,min}}\sum_{(p, \hat{q})\in\mathcal{C}^{(k)}} ||\mathbf{R}p + \mathbf{t} - \hat{q}||^2\;,
\end{equation}
to update the rotation $\mathbf{R}$ and translation $\mathbf{t}$ until convergence.
With this theoretical background, we investigate how ICP contributes to consistent instance ID assignment across frames while enhancing robustness against outliers.

\subsection{Geo-4D}
\label{sec:icp4d}
Our proposed \textsc{Geo-4D} is developed on top of a pre-trained 3D LiDAR panoptic segmentation model~\citep{hong2021lidar, yilmaz2024mask4former}, which is specialized in frame-wise perception. 
Following the standard assumption in LiDAR panoptic segmentation, we transform all points into the world coordinate system using the ego pose.
Through this model, we obtain the per-point semantic and instance prediction $(s_i,\;\ell_i)$ for a given point cloud, where $s_i\in\{1,...,S\}$, $\ell_i\in \mathbb{N}_0$, and $S$ is the number of classes.
Prior to the association stage, we construct a set of matching candidates between $t$ and $t-1$ times by pairing source and destination instances as follows:
\begin{equation}
\mathcal{M}=\mathcal{G}_{\text{src}} \times \mathcal{G}_\text{dst}=\{(a,b) \; | \; a \in \mathcal{G}_\text{src}, \; b \in \mathcal{G}_\text{dst}\},    
\end{equation}
where $\mathcal{G}_\text{src}$ and $\mathcal{G}_\text{dst}$ are respectively denoted as the sets of unique instance IDs in the source and destination. 
Then, we derive the filtered set $\mathcal{M}^{\prime}$ by keeping only the pairs in $\mathcal{M}$ that share the same semantic class.
In this study, we categorize instances into dynamic, static, and missing types to facilitate effective instance association across frames.

\myparagraph{Static Instance.}
\label{sec:sec_static}
Under the assumption that both the center and the shape of a non-moving rigid object remain invariant, we filter static instances based on their centers and further refine them using covariance cues to improve selection fidelity.
Given the matching candidates $\mathcal{M}^{\prime}$, we first represent each instance by the set of 3D points belonging to it. 
For each instance pair $(a_{\text{src}}^{\prime},\;b_{\text{dst}}^{\prime})\in\mathcal{M}^{\prime}$,
we denote the corresponding point sets as $\mathcal{P}(a_{\text{src}}^{\prime})$ and $\mathcal{P}(b_{\text{dst}}^{\prime})$, respectively. Then, we compute the centroid of each point set:
\begin{equation}
\label{eq-static1}
\mu_s
=
\frac{1}{|\mathcal{P}(a^{\prime}_\text{src})|}
\sum_{p \in \mathcal{P}(a^{\prime}_\text{src})} p,
\qquad
\mu_d
=
\frac{1}{|\mathcal{P}(b^{\prime}_\text{dst})|}
\sum_{q \in \mathcal{P}(b^{\prime}_\text{dst})} q,
\end{equation}
where $p, q \in\mathbb{R}^{3}$ represent the world coordinates of points. 
We keep only the pairs satisfying $||\mu_d-\mu_s||_2<\tau_{center}$, where $\tau_{center}$ is the distance threshold for identifying non-moving objects based on center consistency. 
Although this naive approach performs well under ideal conditions, it suffers when applied to noisy predicted point sets, resulting in unsuitable pair associations.
Thus, we introduce an additional covariance-based constraint that measures the similarity of the spatial spread and geometric structure of the instance point sets.
Specifically, the covariance of each point set is computed as follows:
\begin{equation}
\label{eq-cov}
     {\sigma}_{s}=\mathrm{Cov}(\mathcal{P}(a^{\prime}_{\text{src}})),\quad {\sigma}_{d}=\mathrm{Cov}(\mathcal{P}(b^{\prime}_{\text{dst}})). 
\end{equation}
To assess shape consistency, we measure the similarity between the two covariance matrices and use the predefined threshold $\tau_{cov}$ to control the discrepancy between source and destination point set:
\begin{equation}
\label{eq-static2}
     {||{\sigma}_d-{\sigma}_s||_{F}\over \text{tr}({\sigma}_d) + \text{tr}({\sigma}_s)} < \tau_{cov},
\end{equation} 
where $\text{tr}(\cdot)$ and $\|\cdot\|_F$ denote the matrix trace and the Frobenius norm. 
Then, for all candidate pairs satisfying the filtering conditions, we choose the smallest discrepancy in Eq.~\ref{eq-static2} for each source instance to determine the best-matching pairs while avoiding duplicate associations.

\begin{figure}[t]
\begin{center}
   \includegraphics[width=\linewidth]{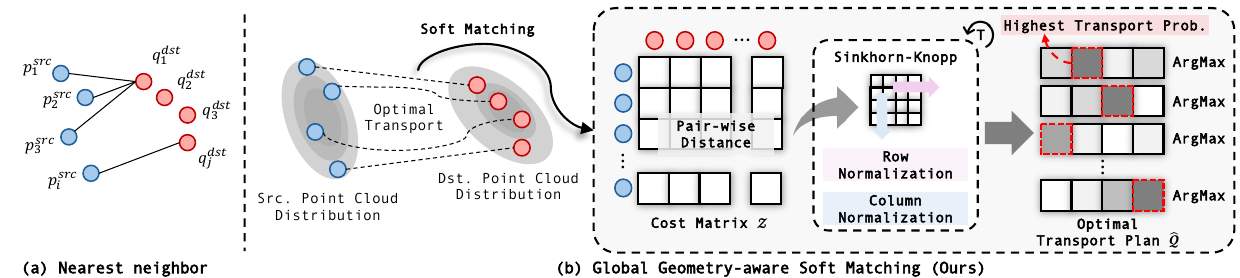}
\end{center}
\vspace{-1em}
\caption{(a) Nearest neighbor aligns each source point with its closest counterpart, focusing only on local point proximity. 
(b) In contrast, our Global-Geometry-aware Soft Matching computes correspondences via an optimal transport plan $\hat{\mathcal{Q}}$, enabling instance-aware matching that respects the global geometry of each instance point set.}
\label{fig:fig_softassign}
\end{figure}

\myparagraph{Dynamic Instance.}
\label{sec:sec_dyn}
After filtering static instances, we construct the remaining candidate pair set $\mathcal{M}^{\prime\prime}$.
These candidate pairs are then fed into the global geometric association,
which serves as the core component of our framework for training-free instance association.
In the vanilla ICP loop, the first step is finding point pairs to compute the distance by utilizing the nearest-neighbor algorithm.
While ICP has a powerful ability to register point cloud sets, determining the corresponding points under noisy instance predictions potentially causes erroneous matches.
Therefore, we propose a Global Geometry-aware Soft Matching that facilitates stable correspondence estimation within ICP. 

\myparagraph{Global Geometry-aware Soft Matching} starts by regarding the point cloud as a probability distribution rather than a discrete set of individual points. 
For a candidate pair $(a^{\prime\prime}_{\text{src}},\;b^{\prime\prime}_{\text{dst}})\in \mathcal{M}^{\prime\prime}$,
we denote their corresponding point sets as $\mathcal{P}(a^{\prime\prime}_{\text{src}}) = \{p_i\}_{0}^{I-1}$ and $\mathcal{P}(b^{\prime\prime}_{\text{dst}}) = \{q_j\}_{0}^{J-1}$, where $I$ and $J$ denote the number of points in each instance. 
We then formulate the correspondence estimation as an Optimal Transport~(OT) problem~\citep{cuturi2013sinkhorn}.
Specifically, we seek a transport plan $\mathcal{Q}$ that minimizes the cost of transforming one distribution into the other as follows:
\begin{equation}
\label{eq-ot}
     \text{OT}(\mathcal{P}(a^{\prime\prime}_\text{src}),\; \mathcal{P}(b^{\prime\prime}_\text{dst})):=\underset{\mathcal{Q}}{\operatorname*{min}}	\langle \mathcal{Z},\;\mathcal{Q} \rangle_F -\epsilon H(\mathcal{Q}),
\end{equation}
where $\langle \cdot \rangle_F$ denotes Frobenius dot-product. Here, the cost matrix $\mathcal{Z}$ represents the pairwise transport cost, defined as $\mathcal{Z}_{ij}=||p_i-q_j||_2^2$. The entropy $H$ and $\epsilon$ control the smoothness of the transport plan as a regularization term.
Then, we utilize the Sinkhorn-Knopp algorithm~\citep{sinkhorn1967diagonal} that provides an efficient solution to the entropy-regularized optimal transport problem by iteratively rescaling the rows and columns of $\mathcal{Q}$ until the marginal constraints are satisfied (details in supplementary material).
Afterwards, we derive the final correspondences from the source to the destination point by taking the $\operatorname{argmax}$ operation over each row of the optimal transport plan $\hat{\mathcal{Q}}$ as:
\begin{equation}
\label{eq-comb_ot}
    j^{*}(i)={\underset{j}{\operatorname*{arg\,max}}}\hat{\mathcal{Q}}_{ij}, \quad \mathcal{C}^{(k)}_\text{GGSM} = \{(p_i,q_{j^{*}(i)}) \mid i = 0,\dots,I-1\}.
\end{equation}
That is, we approximate point-to-point correspondences by selecting the highest transport probability from the minimized transport plan.
As a result, this strategy is based on considering the instance instead of local points~(see Figure~\ref{fig:fig_softassign}), thereby alleviating the vulnerability to outliers and incorrect predictions.

Returning to the ICP procedure, we follow the standard process, initializing the transformation with a histogram-based method~\citep{lin2024icp} to reduce sensitivity to initial estimates and subsequently running iterative refinement to get the final rigid transformation.
After obtaining the transformation ($\mathbf{R}, \mathbf{t}$), we determine whether two instances correspond by evaluating the IoU between their transformed point sets.
Specifically, points with errors smaller than the predefined threshold $\tau_{dist}$ are treated as inliers. Then, in order to filter valid matching candidates, we evaluate alignment accuracy between source and destination instances using IoU scores as follows:
\begin{equation}
\label{eq-iou}
    \delta(a^{\prime\prime}_{\text{src}}, b^{\prime\prime}_{\text{dst}})=\mathds{1} [\text{IoU}(\mathcal{F}(\mathcal{P}(a^{\prime\prime}_{\text{src}})), \mathcal{P}(b^{\prime\prime}_{\text{dst}}))\ge \tau_{\text{iou}}],
\end{equation}
where $\text{IoU}(\cdot)$ denotes the IoU calculation, and $\mathcal{F}$ applies the estimated transformation obtained by ICP.
Using Eq.~\ref{eq-iou}, we assign the same instance ID to matched candidates. In our experiments, this simple matching strategy consistently outperforms more complex assignment schemes. Additional comparisons are provided in the supplementary material.

\myparagraph{Missing Instance.}
Occlusions stemming from the inherent limitations of LiDAR sensor often break temporal continuity beyond two consecutive scans.
To mitigate this, we introduce a memory bank $\mathcal{B}$ that caches the world coordinates, semantic labels, and instance IDs of unmatched objects in both static and dynamic association stages.
Owing to bounded memory capacity, $\mathcal{B}$ keeps only entries observed within a short window of the previous $w_{\text{mem}}$ scans. If $\mathcal{B}$ is not empty, we retrieve the instance information and apply the global geometric association strategy to restore temporal consistency.

%% file: sec/4_experiment.tex
\section{Experiments}
\label{sec:exp}

In this section, we present experimental results to answer the following research questions: \textbf{(Q1)} How well does \textsc{Geo-4D} perform compared with different types of association methods? \textbf{(Q2)} Why are the geometric properties of point clouds essential for reliable instance association? \textbf{(Q3)} Does \textsc{Geo-4D} generalize consistently across different segmentation outputs and datasets? \textbf{(Q4)} How does global geometric consistency help GGSM avoid erroneous local correspondences? \textbf{(Q5)} Does \textsc{Geo-4D} maintain robust performance regardless of object type and point density?

\begin{table*}[t]
\centering
\caption{
\textbf{Quantitative results on SemanticKITTI.}
We compare \textsc{Geo-4D} with prior methods in terms of the standard 4D panoptic metric (LSTQ), complementary tracking metrics, and computational efficiency.
Memory usage is reported in GB, and inference time is reported in seconds per scan.
Best and second best results are shown in \textbf{bold} and \underline{underlined}, respectively.
}
\label{tab:semantickitti_main}

\setlength{\tabcolsep}{4.2pt}
\renewcommand{\arraystretch}{1.06}

\resizebox{\textwidth}{!}{
\begin{tabular}{lc|cccccc|cc|cc}
\toprule
&
&
\multicolumn{6}{c|}{Validation}
&
\multicolumn{2}{c|}{Test}
&
\multicolumn{2}{c}{Efficiency}
\\
\cmidrule(lr){3-8}
\cmidrule(lr){9-10}
\cmidrule(lr){11-12}

Method
& \#Sc.
& LSTQ$\uparrow$
& $S_{\rm assoc}\uparrow$
& MOTSA$\uparrow$
& sMOTSA$\uparrow$
& PTQ$\uparrow$
& sPTQ$\uparrow$
& LSTQ$\uparrow$
& $S_{\rm assoc}\uparrow$
& Mem.$\downarrow$
& Inf. Time$\downarrow$
\\
\midrule

4D-StOP~\citep{kreuzberg20224d}
& 2
& 66.4
& 71.8
& 31.0
& 27.1
& 53.4
& 53.7
& 62.9
& 67.3
& 15.5
& 3.67
\\

Eq-4D-StOP~\citep{zhu20234d}
& 2
& 68.9
& 74.8
& 43.6
& 39.6
& 56.4
& 56.7
& 65.4
& \underline{69.7}
& 11.0
& 2.89
\\

Mask4Former~\citep{yilmaz2024mask4former}
& 2
& \underline{72.0}
& \underline{76.3}
& 47.3
& 42.1
& 48.3
& 48.5
& \underline{68.4}
& 67.3
& 10.4
& 3.51
\\

Mask4D~\citep{marcuzzi2023mask4d}
& 1
& 71.4
& 75.4
& \underline{52.1}
& \underline{48.1}
& \underline{58.4}
& \underline{58.5}
& 64.3
& 66.4
& 11.9
& 1.12
\\

Mask4Former + CA-Net~\citep{marcuzzi2022contrastive}
& 1
& 70.4
& 73.4
& 48.4
& 43.1
& 48.9
& 49.1
& 66.5
& 62.8
& 5.71
& 1.56
\\

\midrule

\rowcolor{gray!12}
\textbf{Mask4Former + \textsc{Geo-4D}}
& \textbf{1}
& \textbf{72.7}
& \textbf{78.3}
& \textbf{73.3}
& \textbf{65.7}
& \textbf{59.6}
& \textbf{59.7}
& \textbf{70.3}
& \textbf{70.0}
& \textbf{3.15}
& \textbf{1.10}
\\

\bottomrule
\end{tabular}
}
\vspace{-4pt}
\end{table*}

\begin{figure*}[htb!]
\begin{center}
   \includegraphics[width=\textwidth]{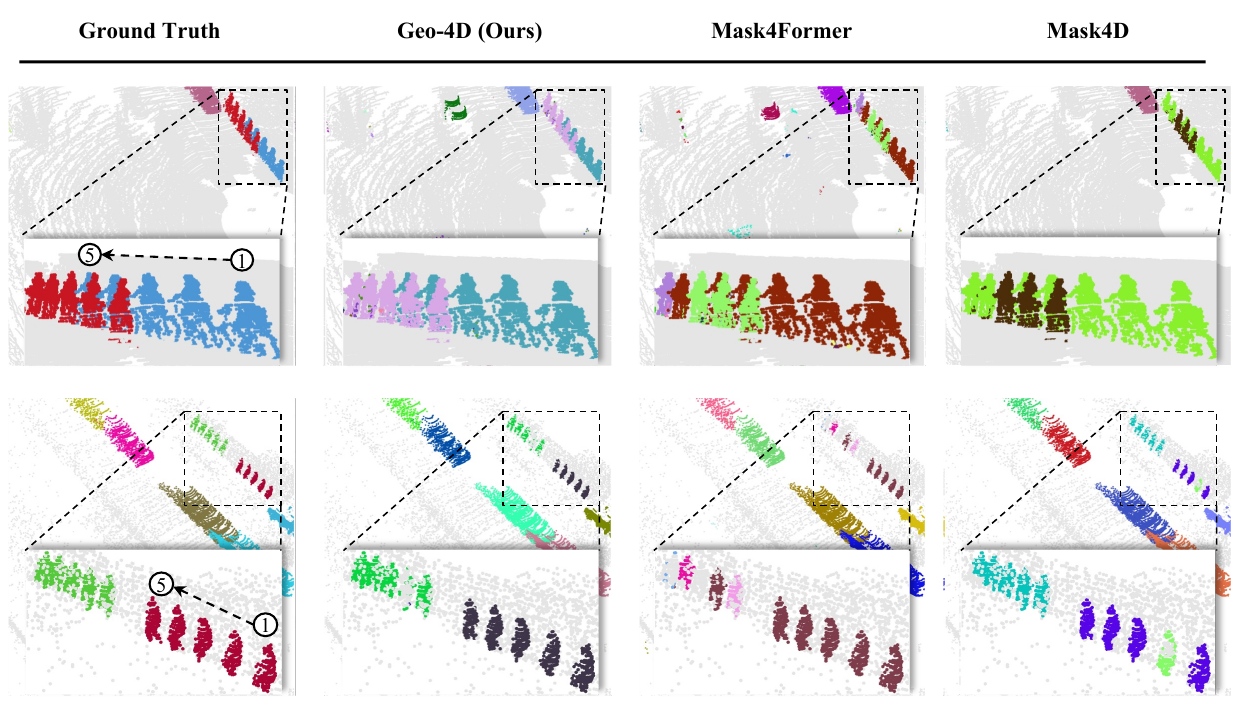}
\end{center}
\vspace{-8pt}
\caption{\textbf{Qualitative comparison on SemanticKITTI}. We visualize the association results over five consecutive scans for both the baselines and our \textsc{Geo-4D}. Different colors represent different instances. The dotted boxes zoom in for clear comparison. For clarity, we use circled numbers (\textcircled{1}–\textcircled{5}) to represent the frame indices.}
\label{fig:fig_qual}
\vspace{-1em}
\end{figure*}

\subsection{Experiment Setup}
\myparagraph{Datasets.}
We choose two representative benchmarks. 
SemanticKITTI~\citep{behley2019semantickitti} captures point clouds using a 64-beam LiDAR at 10 Hz and provides point-wise annotations for 19 semantic classes (8 Things / 11 Stuff).
In contrast, panoptic nuScenes~\citep{fong2022panoptic} uses a 32-beam LiDAR at 2 Hz and offers point-wise annotations for 16 semantic classes (10 Things / 6 Stuff).

\myparagraph{Evaluation Metrics.}
To evaluate tracking quality, we adopt the widely used LSTQ~\citep{Aygun_2021_CVPR}, defined as $\text{LSTQ} = \sqrt{S_{cls} \times S_{assoc}}$.
Here, $S_{cls}$ measures how accurately semantic labels are assigned and $S_{assoc}$ evaluates how consistently instance identities are maintained throughout the sequence, independently of semantic accuracy. A detailed formulation, along with descriptions of the additional evaluation metrics and implementation details, is provided in Appendix~\ref{app:A}.

\subsection{4D LiDAR Panoptic Segmentation Results}

\myparagraph{Addressing Q1.}
Table~\ref{tab:semantickitti_main} reports the experimental results on the SemanticKITTI dataset.
Beyond the standard LSTQ and $S_{\mathrm{assoc}}$ metrics, we report additional tracking metrics to provide a more comprehensive evaluation of instance association quality.
These results reveal two key observations.

First, \textsc{Geo-4D} consistently exhibits strong improvements over competing methods across all tracking metrics, regardless of the number of input scans.
Specifically, on the validation set, \textsc{Geo-4D} improves $S_{\mathrm{assoc}}$ by 2.0 points over the strongest competitor (78.3 vs. 76.3), while achieving substantially larger gains of 21.2 and 17.6 points in MOTSA and sMOTSA, respectively.
Furthermore, \textsc{Geo-4D} maintains strong performance on both the validation and test sets, whereas learning-based methods (Mask4D and CA-Net) exhibit noticeable performance degradation on the test set.
This highlights the robustness of direct geometric association on raw point clouds.

Second, \textsc{Geo-4D} offers the most well-balanced trade-off between performance and computational overhead.
Compared with methods using multiple input scans, \textsc{Geo-4D} requires 69.7\%--73.8\% less memory and reduces inference time by 61.9\%--70.0\%, while achieving superior performance.
Although Mask4D shows competitive runtime, it incurs substantially higher memory overhead due to its transformer-based architecture.
CA-Net is also computationally efficient, but yields considerably weaker association performance, with the gap becoming even more pronounced on the hidden test set.

\begin{table*}[t]
\centering
\caption{
\textbf{Association analysis on SemanticKITTI.}
(a) Comparison with classic association baselines.
(b) Controlled comparison under identical frame-wise predictions.
}
\label{tab:association_analysis}

\begin{minipage}[t]{0.475\textwidth}
\centering
\textbf{(a) Classic association baselines}

\vspace{2pt}
\scriptsize
\setlength{\tabcolsep}{4.5pt}
\renewcommand{\arraystretch}{1.08}

\resizebox{\linewidth}{!}{
\begin{tabular}{lccc}
\toprule
Method
& \#Scan
& LSTQ$\uparrow$
& $S_{\rm assoc}\uparrow$ \\
\midrule

Centroid + Hungarian
& 1 & 55.4 & 45.4 \\

3D Box IoU + Hungarian
& 1 & 63.6 & 59.9 \\

Kalman + 3D Box IoU + Hungarian
& 1 & 64.6 & 61.9 \\

\midrule

\textbf{\textsc{Geo-4D} (Ours)}
& \textbf{1}
& \textbf{72.7}
& \textbf{78.3} \\

\bottomrule

\end{tabular}
}

\vspace{2pt}
{\scriptsize
All baselines use the same Mask4Former predictions.
}
\end{minipage}
\hfill
%
\begin{minipage}[t]{0.505\textwidth}
\centering
\textbf{(b) Same-mask comparison}

\vspace{2pt}
\scriptsize
\setlength{\tabcolsep}{5.5pt}
\renewcommand{\arraystretch}{1.08}

\resizebox{\linewidth}{!}{
\begin{tabular}{llccc}
\toprule
Input Masks
& Association Strategy
& \#Scan
& LSTQ$\uparrow$
& $S_{\rm assoc}\uparrow$ \\
\midrule

\multirow{2}{*}{Eq-4D-StOP}
& IoU-based
& 2
& 65.0
& 71.5 \\

& \textbf{Geo-4D (Ours)}
& 1
& \textbf{65.2}
& \textbf{71.9} \\

\midrule

\multirow{2}{*}{Mask4Former}
& IoU-based
& 2
& 72.0
& 76.3 \\

& \textbf{Geo-4D (Ours)}
& 1
& \textbf{72.8}
& \textbf{77.8} \\

\bottomrule
\end{tabular}
}

\vspace{2pt}
{\scriptsize
Each pair shares identical frame-wise predictions.
}
\end{minipage}
\vspace{-1em}
\end{table*}

\begin{wraptable}{r}{0.48\textwidth}
\vspace{-8pt}
\centering
\caption{
Quantitative results on panoptic nuScenes~\citep{fong2022panoptic}.
}
\label{tab:nuscenes_quan}

\scriptsize
\setlength{\tabcolsep}{3.0pt}
\renewcommand{\arraystretch}{1.05}

\resizebox{\linewidth}{!}{
\begin{tabular}{lcccc}
\toprule
Method
& \#Sc.
& LSTQ$\uparrow$
& $S_{\rm assoc}\uparrow$
& $S_{\rm cls}\uparrow$ \\
\midrule


4D-StOP~\citep{kreuzberg20224d}
 & 4 & 60.5 & 62.5 & 58.6 \\

Eq-4D-StOP~\citep{zhu20234d}
& 2 & 58.9 & 62.3 & 55.7 \\

Mask4Former~\citep{yilmaz2024mask4former}
& 2 & 67.6 & 63.0 & \textbf{72.5} \\

Mask4D~\citep{marcuzzi2023mask4d}
& 1 & 58.5 & 54.1 & 63.1 \\

\midrule

Mask4Former
& 1 & -- & -- & 71.0 \\

\quad + CA-Net~\citep{marcuzzi2022contrastive}
& 1 & 63.4 & 56.6 & 71.0 \\

\quad + \textbf{\textsc{Geo-4D} (Ours)}
& \textbf{1}
& \textbf{67.9}
& \textbf{65.0}
& 71.0 \\

\bottomrule
\end{tabular}
}
\vspace{-4pt}
\end{wraptable}

Moreover, we qualitatively evaluate \textsc{Geo-4D} in Figure~\ref{fig:fig_qual}.
In the first row, \textsc{Geo-4D} consistently preserves the identities of two closely spaced bicyclists, whereas Mask4Former struggles to preserve identity consistency when objects are spatially adjacent or overlapping due to its IoU-based association.
Mask4D also becomes less reliable when objects share similar visual or spatial characteristics, as its association relies on learnable tracking queries.
In the second row, we demonstrate the benefit of our memory bank for temporarily missing instances. Although the 3D panoptic model misses the person shown in green at the second time step, our memory bank successfully retrieves the instance and maintains its identity, while Mask4Former experiences ID switching when it fails to detect an instance due to the absence of overlapping regions.
Consequently, these observations underscore the importance of explicit geometric matching in our approach and a well-structured association pipeline.

\myparagraph{Addressing Q2.}
Table~\ref{tab:association_analysis}(a) highlights the necessity of directly leveraging raw object geometry rather than high-level geometric descriptors~(e.g., centroid and 3D bbox).
The centroid computes the mean of object points and uses Euclidean distance as the matching cost, while the 3D bounding box defines the cost based on 3D IoU. 
We further enhance the baseline with a Kalman filter for temporal state estimation. 
We find that our \textsc{Geo-4D} outperforms classic baselines, yielding 26.5\%--72.5\% improvements in \(S_{\text{assoc}}\).
This finding suggests that exploiting the geometric structure of objects significantly reduces identity ambiguity, resulting in reliable instance association.

\myparagraph{Addressing Q3.}
Table~\ref{tab:association_analysis}(b) evaluates \textsc{Geo-4D} against an IoU-based strategy while holding the segmentation masks fixed. We clearly observe that \textsc{Geo-4D} improves association performance under both mask settings, using predictions from EQ-4D-StOP and Mask4Former. Notably, these gains are achieved even when only single frame segmentation masks are available.
Moreover, Table~\ref{tab:nuscenes_quan} shows the 4D panoptic results on panoptic nuScenes dataset. Consistent with the observations on the SemanticKITTI, \textsc{Geo-4D} delivers notably stronger instance association performance, outperforming existing methods in terms of LSTQ and \(S_{\text{assoc}}\).
Together, these results imply that \textsc{Geo-4D} generalizes across different segmentation masks and datasets, while remaining highly effective under sparse LiDAR observations.

\begin{figure}[t]
\begin{center}
   \includegraphics[width=\linewidth]{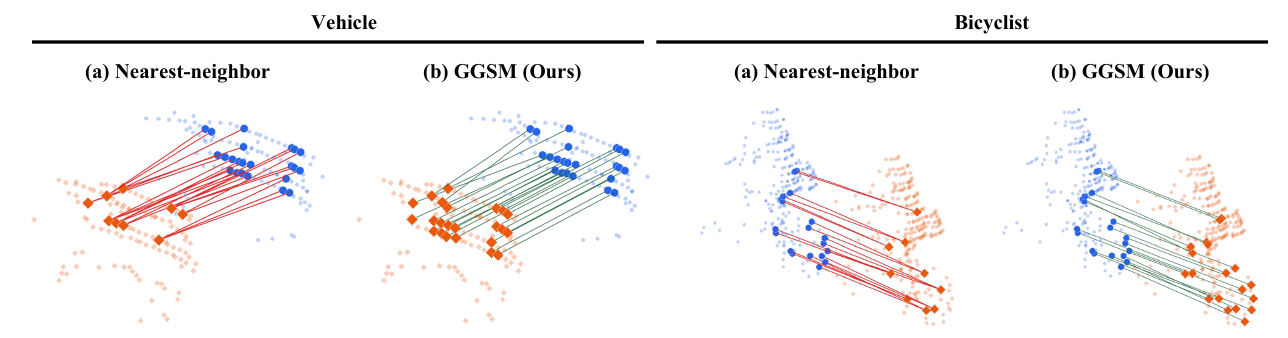}
\end{center}
\vspace{-1em}
\caption{Visualization of point-wise correspondences.
(a) Nearest-neighbor matching.
(b) Our Global Geometry-aware Soft Matching.
\textcolor{orange}{\rotatebox{45}{\scalebox{0.5}{$\blacksquare$}}} and $\textcolor{blue}{\bullet}$ represent the target and source points, respectively.
}
\label{fig:fig_comp_sink}
\vspace{-0.5em}
\end{figure}

\begin{figure}[t]
\begin{center}
   \includegraphics[width=\linewidth]{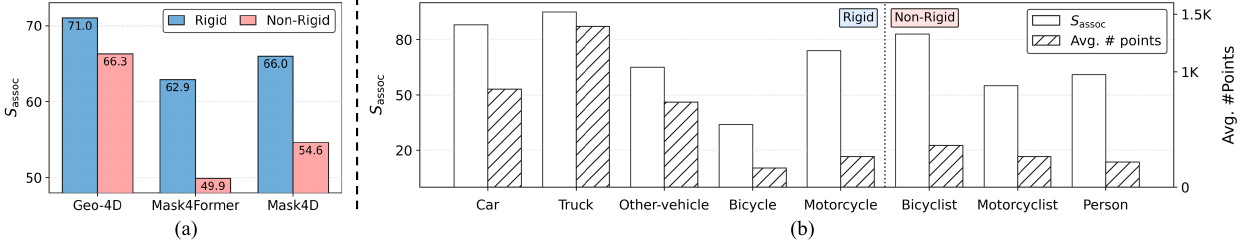}
\end{center}
\vspace{-8pt}
\caption{(a) Comparison with representative baselines across different association strategy categories on rigid and non-rigid object classes. (b) Class-wise performance with respect to object class types and the average number of points per instance.}
\label{fig:fig_comp_class}
\vspace{-8pt}
\end{figure}

\subsection{Further Analysis of the Global Geometric Association}

\myparagraph{Addressing Q4.}
The Global Geometry-aware Soft Matching~(GGSM) forms instance-aware point correspondence between source and target points.
Owing to this strategy, accurate error measurements can be obtained even in the presence of outliers, as the estimation is not dominated by local proximity.
Figure~\ref{fig:fig_comp_sink} visualizes the point-wise correspondences obtained from two consecutive scans for 20 randomly selected points, where the current scan is used as the source and the previous scan as the target.
For this analysis, we extract each point segment at the initial stage of ICP, as this stage retains its geometric structure intact.
We observe that the nearest-neighbor matching is biased toward the closest point, ignoring the underlying point-wise relationships of the instance~(\textcolor{red}{red} lines).
In contrast, applying GGSM on the baseline maintains the geometric structure, yielding correspondences that align consistently with the instance's motion direction~(\textcolor{ForestGreen}{green} lines).

\myparagraph{Addressing Q5.}
We categorize object classes into rigid (car, truck, other-vehicle, bicycle, motorcycle) and non-rigid (bicyclist, motorcyclist, person) groups based on their physical characteristics.
As shown in Figure~\ref{fig:fig_comp_class}(a), \textsc{Geo}-4D outperforms representative IoU-based and query-based association baselines across both rigid and non-rigid classes.
Furthermore, \textsc{Geo}-4D consistently maintains strong association performance across all object classes (see Figure~\ref{fig:fig_comp_class}(b)).
These results are attributed to the robustness of our proposed global geometry-aware association strategy, which captures holistic geometric structure and enables reliable association under partial non-rigid motion and significant variations in object size.

\subsection{Ablation Study}
We conduct a series of ablation studies to validate the effectiveness of each component in our framework.
All experiments are performed on the SemanticKITTI~\citep{behley2019semantickitti} validation set, with \textsc{Geo-4D} integrated into Mask4Former~\citep{yilmaz2024mask4former}.

\myparagraph{Module Design.}
Table~\ref{tab:ablation_main}(a) shows the effects of our proposed methods.
Exp. \#1 directly applies the vanilla Iterative Closest Point~(ICP) for associating instances, serving as a baseline to examine the feasibility of ICP in an unsupervised manner. 
Despite its strong matching capability, ICP is highly sensitive to the quality of initial alignment. In Exp. \#2, we observe a little performance gain when applying histogram-based initialization proposed by~\citep{lin2024icp}, which stabilizes the initial alignment.
Notably, our Global Geometry-aware Soft Matching~(GGSM) brings further improvements without the influence of initial pose (see Exp. \#3).
The key takeaway is that instance-aware correspondence to alleviate the instability of local point matching is highly beneficial for obtaining reliable matching points between noisy instance segments, and as a result helps to estimate an accurate transformation.
Exp. \#4 demonstrates that coupling these components yields a complementary effect, leading to more robust instance association and improved overall performance.
Lastly, Exp. \#5 demonstrates the effectiveness of incorporating a memory bank for handling occluded and missing instances, resulting in superior performance.

\myparagraph{Effect of Different Types of Instances.}
To validate our carefully designed pipeline with respect to different instance types, Table~\ref{tab:ablation_main}(b) presents the association results and average inference time per a single scan.
First, we employ a statistics-based geometric filtering for static instances to match candidates over all detected objects.
Despite requiring minimal computation, only a small subset of instances remain consistently tracked, 18.5 percent points indicating that static cues alone are insufficient for reliable association.
In contrast, relying solely on global geometry association achieves high performance, but is computationally expensive.
Thus, combining these strategies reduces the runtime by 48.4\% while simultaneously improving performance, as it significantly narrows the matching candidates for global geometry association.
Consequently, our full \textsc{Geo-4D} enhances the tracking continuity without sacrificing computational efficiency.

\begin{table*}[t]
\centering
\caption{
(a) Effect of each module in \textsc{Geo-4D}.
(b) Effect of handling different instance types.
}
\label{tab:ablation_main}
\vspace{-4pt}
\begin{minipage}[t]{0.48\textwidth}
\centering
\textbf{(a) Module design}

\vspace{2pt}
\scriptsize
\setlength{\tabcolsep}{5.0pt}
\renewcommand{\arraystretch}{1.07}

\resizebox{\linewidth}{!}{
\begin{tabular}{cccccc}
\toprule
& \multicolumn{3}{c}{Components}
& \multicolumn{2}{c}{Metric} \\
\cmidrule(lr){2-4}
\cmidrule(lr){5-6}

Exp.
& Hist. Init.
& GGSM
& Memory
& LSTQ$\uparrow$
& $S_{\rm assoc}\uparrow$ \\
\midrule

1
& -- & -- & --
& 70.1 & 73.4 \\

2
& \cmark & -- & --
& 71.3 & 75.4 \\

3
& -- & \cmark & --
& 71.4 & 75.6 \\

4
& \cmark & \cmark & --
& 72.0 & 76.8 \\

5
& \cmark & \cmark & \cmark
& \textbf{72.7}
& \textbf{78.3} \\

\bottomrule
\end{tabular}
}
\end{minipage}
\hfill
%
\begin{minipage}[t]{0.49\textwidth}
\centering
\textbf{(b) Instance-type handling}

\vspace{2pt}
\scriptsize
\setlength{\tabcolsep}{3.0pt}
\renewcommand{\arraystretch}{1.07}

\resizebox{\linewidth}{!}{
\begin{tabular}{cccccc}
\toprule
\multicolumn{3}{c}{Instance Type}
& \multicolumn{3}{c}{Metric} \\
\cmidrule(lr){1-3}
\cmidrule(lr){4-6}

Static
& Dynamic
& Missing
& LSTQ$\uparrow$
& $S_{\rm assoc}\uparrow$
& Time (s)$\downarrow$ \\
\midrule

\cmark & -- & --
& 35.4 & 18.5 & 0.82 \\

-- & \cmark & --
& 71.8 & 76.7 & 1.86 \\

\cmark & \cmark & --
& 72.0 & 76.8 & 0.96 \\

\cmark & \cmark & \cmark
& \textbf{72.7}
& \textbf{78.3}
& 1.10 \\

\bottomrule
\end{tabular}
}
\end{minipage}
\end{table*}

\begin{figure*}[t!]
\begin{center}
   \includegraphics[width=\textwidth]{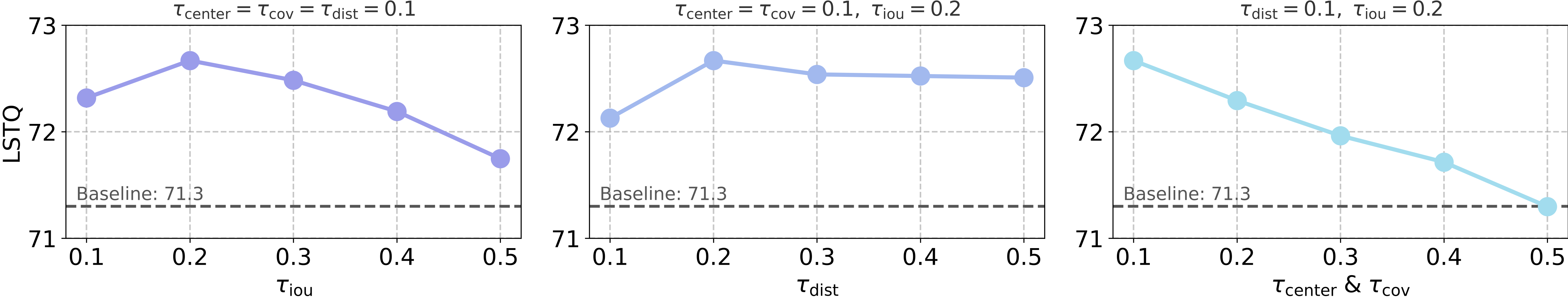}
\end{center}
\vspace{-6pt}
\caption{
We illustrate the performance variation w.r.t. the hyperparameters of \textsc{Geo-4D}.
The dashed line denotes the average performance of three different association methods:
Mask4Former~\citep{yilmaz2024mask4former},
Mask4D~\citep{marcuzzi2023mask4d},
and CA-Net~\citep{marcuzzi2022contrastive}.
}
\label{fig:fig_hyper}
\vspace{-1em}
\end{figure*}

\myparagraph{Hyperparameter Analysis}
\label{app:C3}
In this section, we examine the impact of different matching thresholds on instance association performance~(see Figure~\ref{fig:fig_hyper}).
We empirically determine the matching thresholds \{$\tau_{iou}$, $\tau_{dist}$, $\tau_{center}$, $\tau_{cov}$\} over \{0.1, 0.2, 0.3, 0.4, 0.5\}.
Here, we observe three insights. 
First, low value of threshold $\tau_{iou}$ often yields decent performance due to over-association, whereas high thresholds lead to over-filtering and degraded matching quality. 
Second, large $\tau_{center}$ and $\tau_{cov}$ cause static instances to be matched too liberally, leading to unreliable association.
Lastly, our proposed \textsc{Geo-4D} delivers performance gains over the baselines, regardless of the hyperparameter choices.

%% file: sec/5_conclusion.tex
\vspace{-0.5em}
\section{Conclusion}
\vspace{-0.5em}
\label{sec:conclusion}
In this paper, we introduce a novel instance association strategy \textsc{Geo-4D} for 4D LiDAR panoptic segmentation, the first training-free framework tailored for a single scan setting. 
Our proposed Global Geometry-aware Soft Matching forming instance-aware correspondence enables ICP procedure to precisely align between two consecutive instance point segments, thereby improving temporal continuity.
Additionally, we systematically design the overall pipeline, underscoring the instance types to alleviate the inherent challenges of ICP and LiDAR sensors. 
Through experimental evaluations on both SemanticKITTI and nuScenes, our \textsc{Geo-4D} showcases its superiority not only on the validation sets but also on the hidden test sets, demonstrating robustness across various sensor configurations.
Ultimately, our approach offers a promising direction for 4D perception by solely leveraging strong 3D panoptic models to support temporal understanding.
\myparagraph{Limitations.}
Although \textsc{Geo-4D} shows substantial promise in advancing 4D perception, several limitations persist that warrant further investigation, particularly in real-time deployment, handling partial LiDAR observations, and dependence on accurate ego pose estimation.
We provide additional discussions in the supplementary materials.

%% file: sec/X_suppl.tex
\appendix


\setcounter{section}{0}
\def\thesection{\Alph{section}}
\setcounter{table}{0}
\setcounter{figure}{0}
\renewcommand{\thetable}{A.\arabic{table}}
\renewcommand{\thefigure}{A.\arabic{figure}}

\section{Implementation Details}
\label{app:A}

\subsection{Baseline Details}
\label{app:A1}
In this section, we provide a detailed overview of the strong baselines considered in our comparison: 4D-PLS~\citep{Aygun_2021_CVPR}~(CVPR'21), 4D-Stop (ECCVW'22)~\citep{kreuzberg20224d}, EQ-4D-Stop~(ICCV'23)~\citep{zhu20234d}, CA-Net~(R-AL'22)~\citep{marcuzzi2022contrastive}, Mask4D~(R-AL'23)~\citep{marcuzzi2023mask4d}, and Mask4Former~(ICRA'24)~\citep{yilmaz2024mask4former}.
Note that we report the official results as stated in the original papers, and reproduce the results when formal reports are not available.
The following paragraph details the experimental settings and notable modifications for each baseline.

First, 4D-PLS, 4D-Stop, and EQ-4D-Stop follow a two-stage design, where instance tracklets are first generated independently for all scans and then stitched across consecutive frames.
Accordingly, we measure the inference time of each stage separately and report their average as the overall runtime.
Second, Mask4Former is originally developed for 4D LiDAR panoptic segmentation with superimposed point clouds.
To establish a strong baseline for our experiments, we train it as a 3D panoptic network without any modifications, except setting the number of input scans to one.
Third, CA-Net employs contrastive learning for instance association on top of an off-the-shelf 3D panoptic segmentation network.
For a fair comparison, we integrate the single scan Mask4Former into the CA-Net framework.
Since CA-Net requires point-wise instance features for contrastive learning, we extract them by computing the dot product between instance queries and instance masks.
Lastly, we additionally reproduce the baselines on panoptic nuScenes~\cite{fong2022panoptic} using their official repositories, following the experimental protocol of EQ-4D-StOP.

\subsection{Metrics}
\label{app:A2}
In this section, we provide detailed explanations of the metrics employed for evaluating association performance in our experiments. While LSTQ is the most widely used metric, we additionally report two complementary metrics, MOTSA and PTQ, to provide a more comprehensive assessment.

\myparagraph{LSTQ}~\citep{Aygun_2021_CVPR} is delicately designed to evaluate 4D panoptic segmentation quality.
First, to assess the semantic segmentation performance over the classes $C$, the classification score S$_{cls}$ is defined as:

\begin{equation}
\label{eq:lstq}
     S_{cls}={1\over{|C|}}\sum_{c=1}^{C} {|\text{TP}_{c}|\over {|\text{TP}_{c}|+|\text{FN}_{c}|+|\text{FP}_{c}|}} = {1\over{C}}\sum_{c=1}^{C}IoU(c),
\end{equation}
where TP, FN, and FP denote the true positives, false negatives, and false positives, respectively.
Furthermore, the association score S$_{assoc}$, which is responsible for measuring temporal continuity, is defined as:

\begin{equation}
\label{eq:assoc}
     S_{assoc}={1\over{|I|}}\sum_{t\in I}{1\over |{gt_{id}(t)}|}\sum_{s\in S,\;s\cap t\ne0} TPA(s,t)IoU(s, t),
\end{equation}
where $I$ and $S$ refer to the ground-truth and predicted segments. Note that $TPA$ represents the points that maintain consistent identities across scans, regardless of class.
Consequently, LSTQ is computed as the geometric mean of the classification score S$_{cls}$ and association score S$_{assoc}$.

\myparagraph{MOTSA \& sMOTSA}~\citep{voigtlaender2019mots} are mask-based variants of the widely used Multi-Object Tracking Accuracy (MOTA) metric, replacing bounding-box overlaps with mask IoU to better evaluate segmentation-level tracking performance.
Formally, 
\begin{equation}
\label{eq:motsa}
     \text{MOTSA} = {{|TP|-|FP|-|IDSW|}\over{|M|}},
\end{equation}
where $|IDSW|$ denotes the total number of ID switches, and $|M|$ is the number of ground truth segmentation masks.
In addition, sMOTSA replaces the true positive~(${TP}$) with soft true positive ($sTP$), obtained by accumulating the mask IoUs over all matched predictions.

\myparagraph{PTQ \& sPTQ}~\citep{hurtado2020mopt} are formulated to satisfy the fundamental requirement that panoptic tracking metrics must jointly capture segmentation accuracy and tracking consistency over time.
Thus, they incorporate an ID-switch penalty into the original Panoptic Quality (PQ) measure, as follows:
\begin{equation}
\label{eq:ptq}
     \text{PTQ} = {1\over{|C|}}\sum_{c=1}^C {\sum_{(s, t)\in TP_c}IoU(s,t)-{|IDSW_c|}\over {|TP_c|+{1\over2}|FP_c|+{1\over2}|FN_c|}}.
\end{equation}
Furthermore, sPTQ leverages the soft version of IDSW (sIDSW) instead of IDSW. This soft variant aggregates the IoU scores for segments in which the predicted instance ID $ID$ changes:
\begin{equation}
\label{eq:sidsw}
     sIDSW_c = \{IoU(s,\;t) \;\mid\; (s,\;t)\in TP_c\land ID_s\neq ID_t \}.
\end{equation}

\begin{algorithm}[t]
    \caption{Overall process of \textsc{Geo-4D}}\label{algorithm:icp4d}
    \begin{algorithmic}[1]
        \REQUIRE Panoptic prediction $(s, l)$, memory bank $\mathcal{B}$
        \FOR{sequence $t \in \{1, 2, \dots, T-1\}$}
        \STATE Construct the candidate set $\mathcal{M^\prime} = \mathcal{G}'_{\text{src}} \times \mathcal{G}'_{\text{dst}}$ filtered by semantic predictions $s$.

        \textbf{\textcolor{blue!50}{// Static instances}}
        \STATE Associate instances using the centroid $\mu$ of each point set (Eq.~5).
        \STATE Refine the matched pairs using the covariance descriptor $\sigma$ (Eq.~7).
        
        \textbf{\textcolor{blue!50}{// Dynamic instances}}
        \STATE Perform global geometric association using ICP enhanced with global geometry-aware soft matching.
        
        \textbf{\textcolor{blue!50}{// Missing instances}}
        \IF{$\mathcal{B}$ is not empty}
        \STATE Apply the dynamic matching procedure (Lines~5) to missing instances.
        \ENDIF

        \STATE Update the memory bank $\mathcal{B}$.
        \ENDFOR
    \ENSURE Consistent instance IDs across the entire sequence.
    \end{algorithmic}
\end{algorithm}

\section{Method Details}
\label{app:B}

\subsection{Pipeline Details}
\label{app:B1}
The overall association pipeline of our proposed \textsc{Geo-4D} is summarized in Algorithm~\ref{algorithm:icp4d}.
We apply \textsc{Geo-4D} sequentially across all scans within each sequence, and reset the memory bank $\mathcal{B}$ at the beginning of every new sequence.

\subsection{Complexity Analysis}
\label{app:B2}
In this section, we provide a computational complexity comparison between Global Geometry-aware Soft Matching~(GGSM) and nearest neighbor-based matching.
As described in Section.3.2, we utilize the entropy-regularized optimal transport problem~\citep{cuturi2013sinkhorn} to obtain instance-aware point correspondences.
Before analyzing the regularized form, we briefly review the optimal transport problem.
Given two probability vectors $r\in\mathbb{R}_+^{n}\;\text{and}\;c\in\mathbb{R}_+^{m} $, the original optimal transport problem~\citep{villani2008optimal} aims to find the optimal transport plan $\mathcal{Q}\in U(r,\;c)$. 
The entries of $\mathcal{Q}$ represent the amount of mass transported from the $i$-th element of $r$ to the $j$-th element of $c$.
The transport polytope $U$ is defined as:
\begin{equation}
\label{eq:og_ot_polytope}
     U(r,\;c) =
     \{Q\in\mathbb{R}^{n\times m}_+|\sum_{j}Q_{ij}=r_i,\; \forall i,\; \sum_i Q_{ij}=c_j,\;\forall j\}.
\end{equation}
Let $\mathcal{Z}$ denote the transport cost matrix. The optimal transport plan is then computed as the Frobenius inner product between $\mathcal{Q}$ and $\mathcal{Z}$:

\begin{equation}
\label{eq:og_ot}
     \text{OT}(r,c):=\underset{\mathcal{Q}}{\operatorname*{min}}	\langle \mathcal{Z},\;\mathcal{Q} \rangle_F.
\end{equation}
This optimization is a linear program, for which the network simplex~\citep{bazaraa2011linear} is the most widely used exact solver.
In the symmetric case $n=m$, its worst-case computational complexity scales as $O(n^3\log n)$.
However, due to its heavy computational cost, adding an entropic regularization is widely adopted to accelerate optimal transport.
By incorporating the Shannon entropy $H(\mathcal{Q}):=-\sum_{i,j}\mathcal{Q}_{ij}\log \mathcal{Q}_{ij}$ into Eq.~\ref{eq:og_ot}, we obtain the entropy-regularized optimal transport formulation.
To handle marginal constraints, we introduce dual variables $f\in\mathbb{R}^n$ and $g\in\mathbb{R}^m$ and form the corresponding Lagrangian.
Then, by taking the first-order optimality condition, we find that optimal solution $\hat{\mathcal{Q}}$ has matrix form $\hat{\mathcal{Q}}=\text{diag}(u)K \text{diag}(v)$, where $u_i=e^{f_i/\epsilon}$, $v_j=e^{g_j/\epsilon}$, and $K_{ij}=e^{-Z_{ij}/\epsilon}$.
Since $u$ and $v$ must satisfy the marginal constraints, the entropic regularization transforms the classical optimal transport problem from a linear program into a matrix scaling problem. 
Thus, when $n=m$, the computational complexity is reduced to $O(n^2)$.
However, the nearest neighbor algorithm operates in $O(n\log n)$ on average and up to $O(n^2)$ in the worst case in practice, as implemented in the scikit-learn library~(e.g., using KD-Tree~\citep{bentley1975multidimensional}).
Consequently, our GGSM solution is theoretically comparable to the nearest neighbor algorithm.

\subsection{Option: Bijective assignment.}
In global geometric association, a naive approach is to pair each source instance with its most similar target instance; however, it fails to enforce one-to-one correspondence.
Thus, we derive a transformation-aware matching cost that integrates the ICP results—rotation $\mathbf{R}$, translation $\mathbf{t}$, and the IoU score $s_{\text{iou}}$.
Formally, the matching cost $\mathcal{C}$ of dynamic matching pairs $(i, j)\in \mathcal{M}^{\prime\prime}$ is defined as:

\begin{equation}
\label{eq-cost_eq}
\begin{aligned}
    \mathcal{C}_{ij} &= \gamma_{t} c_{ij}^{(t)} + \gamma_{r}\, c_{ij}^{(r)}+ \gamma_{s}\, c_{ij}^{(s)}, \\
    c_{ij}^{(t)} = \|\mathbf{t}_{ij}\|_2,\quad 
    &c_{ij}^{(r)} = \| \theta_{ij}\|_2,\quad 
    c_{ij}^{(s)} = \|1-s_{iou}\|_2,
\end{aligned}
\end{equation}
where $\theta$ denotes the axis–angle rotation vector obtained from $\mathbf{R}$.
We normalize each cost term to comparable scales before weighting so that $\gamma_{t}$, $\gamma_{r}$, and $\gamma_{s}$ control their relative contributions.
By minimizing $\mathcal{C}$ with the Hungarian algorithm~\citep{kuhn1955hungarian}, we obtain a bijective assignment that considers the geometric alignment.

\subsection{Leaderboard Results}
\label{app:C1}
Table~\ref{table:leader} shows the SemanticKITTI~\citep{behley2019semantickitti} challenge leaderboard ranks on the hidden test set, ordered by LSTQ.
In this leaderboard, we observe that our proposed \textsc{Geo-4D} outperforms not only all published methods but also every competition participant.
Specifically, we achieve 1\textsuperscript{st} place in LSTQ as of submission date.
These results demonstrate the strong generalization capability and practical effectiveness of our method under challenging scenarios.


\begin{table}[t]
\centering
\footnotesize
\caption{SemanticKITTI~\citep{behley2019semantickitti} Test Set Leaderboard (Top 5). 
mAQ and mIoU denote S$_{assoc}$ and S$_{cls}$, respectively.}

\begin{tabular}{ccccc}
\toprule
\textbf{Rank} & \textbf{User} & \textbf{LSTQ} $\uparrow$ & \textbf{mAQ} $\uparrow$ & \textbf{mIoU} $\uparrow$ \\
\midrule
\rowcolor{green!10}1 & \textbf{\textsc{Geo-4D} (Ours)} & \textbf{70.3} & 70.0 & \textbf{70.5} \\
2 & guysh         & 70.2 & 70.6 & 69.7 \\
2 & YAKD         & 68.6 & \textbf{70.8} & 66.5 \\
3 & hphnngcquan  & 68.5 & 70.4 & 66.7 \\
4 & KadirYilmaz  & 68.4 & 67.3 & 69.6 \\
\bottomrule
\end{tabular}
\label{table:leader}
\end{table}

\section{Additional Experimental Results}
\label{app:C}

\subsection{Analysis of Matching Strategy.}
\label{exp:abl_hung}
\begin{wraptable}{r}{0.48\textwidth}
\vspace{-10pt}
\centering
\caption{Ablation on matching strategy.}
\label{tab:matching}
\setlength{\tabcolsep}{5pt}
\begin{tabular}{lccc}
\toprule
Method & LSTQ & S$_{\text{assoc}}$ & S$_{\text{cls}}$ \\
\midrule
Hungarian Matching & 72.3 & 77.6 & 67.5 \\
Greedy Matching    & 72.7 & 78.3 & 67.5 \\
\bottomrule
\end{tabular}
\vspace{-8pt}
\end{wraptable}
In Table~\ref{tab:matching}, we compare the performance of \textsc{Geo-4D} with one-to-one or greedy matching.
The evaluation is done using SemanticKITTI validation set and set all $\gamma_{\{t, r, s\}} = 1$ for the Hungarian matching.
We observe that greedy matching surpasses the bijective matching by a small margin in S$_{\text{assoc}}$. 
We interpret this gap as being caused by segmentation noise, where a single object is often predicted as multiple disjoint segments.
In such case, the greedy strategy adapts to this variability, resulting in more robust temporal consistency.
However, as segmentation quality improves, bijective matching becomes advantageous for maintaining unique and consistent identities.

\subsection{Effect of the Number of Scans.}
\label{app:C2}

In Table~\ref{tab:a1_semantickitti_per_scan}, we examine how well our proposed \textsc{Geo-4D} performs instance association with only a single scan.
The number of superimposed scans is a pivotal factor influencing the association performance of the IoU-based method.
Since formal reports for specific configurations are unavailable, we reproduce the results using official implementation—namely, Eq-4D-Stop~\citep{zhu20234d} with 2 scans and Mask4Former~\citep{yilmaz2024mask4former} with 4 scans.
As clearly observed in Table~\ref{tab:a1_semantickitti_per_scan}, leveraging four superimposed scans yields a substantial improvement in S$_\text{assoc}$, with a notable sacrifice in computational efficiency.
In contrast, we observe that \textsc{Geo-4D} exhibits superior performance, despite the absence of explicit 4D spatio-temporal volume constructed from stacked point clouds.
This observation demonstrates that \textsc{Geo-4D} effectively associates instances using only a single scan and a 3D panoptic network, while remaining highly efficient.

\begin{table*}[t!]
\centering
\footnotesize
\caption{Analysis on performance over the number of superimposed scans.}
\label{tab:a1_semantickitti_per_scan}

\setlength{\tabcolsep}{4pt}  
\renewcommand{\arraystretch}{1.15} 

\resizebox{\textwidth}{!}{
\begin{tabular}{lcccccccc ccccc}
\toprule
 & & \multirow[b]{2}{*}{\raisebox{-0.3\height}{\shortstack{Memory\\ (GB)}}}
 & \multirow[b]{2}{*}{\raisebox{-0.3\height}{\shortstack{Time\\ (s)}}}
 & \multicolumn{5}{c}{Validation set} 
 & \multicolumn{5}{c}{Test set} \\
 \cmidrule(lr){5-9} \cmidrule(lr){10-14}
 Method & \# Scan & & & LSTQ & S$_\text{assoc}$ & S$_\text{cls}$ & IoU$^{St}$ & IoU$^{Th}$ 
        & LSTQ & S$_\text{assoc}$ & S$_\text{cls}$ & IoU$^{St}$ & IoU$^{Th}$ \\
\midrule

\multicolumn{13}{l}{\textbf{\textit{(a) IoU-based Association}}} \\

\multirow{2}{*}{4D-PLS~\citep{Aygun_2021_CVPR}} 
 & 2 & - & - & 59.9 & 58.8 & 61.0 & 65.0 & 63.1 & - & - & - & - & - \\
 & 4 & 8.0 & 3.48 & 62.7 & 65.1 & 60.5 & 65.4 & 61.3 & 56.9 & 56.4 & 57.4 & 66.9 & 51.6 \\
\arrayrulecolor{black!10}\midrule\arrayrulecolor{black}

\multirow{2}{*}{4D-Stop~\citep{kreuzberg20224d}} 
 & 2 & 15.9 & 3.67 & 66.4 & 71.8 & 61.4 & 64.9 & 64.1 & 62.9 & 67.3 & 58.8 & 68.3 & 53.3 \\
 & 4 & 18.9 & 4.32 & 67.0 & 74.4 & 60.3 & 65.3 & 60.9 & 63.9 & 69.5 & 58.8 & 67.7 & 53.8 \\
\arrayrulecolor{black!10}\midrule\arrayrulecolor{black}

\multirow{2}{*}{Eq-4D-Stop~\citep{zhu20234d}}
 & 2 & 11.3 & 2.89 & 68.9 & 74.8 & 63.5 & 65.7 & 68.4 & 65.4 & 69.7 & 61.5 & 68.7 & 59.1 \\
 & 4 & 12.1 & 3.18 & 70.1 & \underline{77.6} & 63.4 & 66.4 & 67.1 
       & 67.0 & \textbf{72.0} & 62.4 & 69.1 & 60.9 \\
\arrayrulecolor{black!10}\midrule\arrayrulecolor{black}

\multirow{2}{*}{Mask4Former~\citep{yilmaz2024mask4former}}
 & 2 & 10.4 & 3.51 & \underline{72.0} & 76.3 & 68.0 & 66.2 & 70.5 
       & \underline{68.4} & 67.3 & 69.6 & 72.7 & 65.3 \\
 & 4 & 17.1 & 8.56 & 71.9 & 76.3 & 67.8 & 66.9 & 69.0 
       & 68.0 & 69.7 & 66.3 & 70.7 & 60.3 \\
\arrayrulecolor{black!10}\midrule\arrayrulecolor{black}

\multicolumn{13}{l}{\textbf{\textit{(c) Detect \& Track Association}}} \\
\rowcolor{gray!20}
\textbf{\textsc{Geo-4D} (Ours)}  
 & 1 & \textbf{3.2} & \textbf{1.10} & \textbf{72.7} & \textbf{78.3} & 67.5 & 65.2 & 70.6 
       & \textbf{70.3} & \underline{70.0} & 70.5 & 72.7 & 67.4 \\

\bottomrule
\end{tabular}
}
\end{table*}

\begin{table*}[t!]
\centering
\caption{Comparison of Geo-4D with CA-Net using different frame-wise segmentation backbones.}
\label{tab:detect_and_track}
\setlength{\tabcolsep}{1pt}
\resizebox{\textwidth}{!}{
\begin{tabular}{lccccccc ccccc}
\toprule
& & \multirow[b]{2}{*}{\raisebox{-0.3\height}{\shortstack{Assoc. \\ Training}}}
& \multicolumn{5}{c}{Validation set}
& \multicolumn{5}{c}{Test set} \\
\cmidrule(lr){4-8} \cmidrule(lr){9-13}

Method & \# Scan & 
& LSTQ & S$_\text{assoc}$ & S$_\text{cls}$ & IoU$^{St}$ & IoU$^{Th}$
& LSTQ & S$_\text{assoc}$ & S$_\text{cls}$ & IoU$^{St}$ & IoU$^{Th}$ \\
\midrule

DS-Net~\citep{hong2021lidar}
& 1 & -
& - & - & 63.5 & 64.6 & 62.0
& - & - & 60.6 & 66.9 & 52.0 \\

\hspace{0.5em}$\boldsymbol{+}$ CA-Net~\citep{marcuzzi2022contrastive}
& 1 & \cmark
& 67.4 & 71.6 & 63.5 & 64.6 & 62.0
& 62.1 & 63.7 & 60.6 & 66.9 & 52.0 \\

\rowcolor{gray!20}
\hspace{0.5em}$\boldsymbol{+}$ \textbf{\textsc{Geo-4D} \textit{(Ours)}}
& 1 & \xmark
& 68.5~{\footnotesize\textcolor{blue}{(+1.1)}}
& 74.0~{\footnotesize\textcolor{blue}{(+2.4)}}
& 63.5 & 64.6 & 62.0
& 62.9~{\footnotesize\textcolor{blue}{(+0.8)}}
& 65.3~{\footnotesize\textcolor{blue}{(+1.6)}}
& 60.6 & 66.9 & 52.0 \\

\arrayrulecolor{black!10}\midrule\arrayrulecolor{black}

Mask4Former$^*$~\citep{yilmaz2024mask4former}
& 1 & -
& - & - & 67.5 & 65.2 & 70.6
& - & - & 70.5 & 72.7 & 67.4 \\

\hspace{0.5em}$\boldsymbol{+}$ CA-Net$^*$
& 1 & \cmark
& 70.4 & 73.4 & 67.5 & 65.2 & 70.6
& 66.5 & 62.8 & 70.5 & 72.7 & 67.4 \\

\rowcolor{gray!20}
\hspace{0.5em}$\boldsymbol{+}$ \textbf{\textsc{Geo-4D} \textit{(Ours)}}
& 1 & \xmark
& \textbf{72.7}~{\footnotesize\textcolor{blue}{(+2.3)}}
& \textbf{78.3}~{\footnotesize\textcolor{blue}{(+4.9)}}
& 67.5 & 65.2 & 70.6
& \textbf{70.3}~{\footnotesize\textcolor{blue}{(+3.8)}}
& \textbf{70.0}~{\footnotesize\textcolor{blue}{(+7.2)}}
& 70.5 & 72.7 & 67.4 \\

\bottomrule
\end{tabular}
}
\end{table*}

\subsection{Controlled same-mask comparison with Detect \& Track association.}
We further conduct a controlled same-mask comparison with CA-Net~\citep{marcuzzi2022contrastive} on SemanticKITTI dataset~\citep{behley2019semantickitti}.
For each segmentation backbone, we fix the frame-wise predictions and apply either CA-Net or Geo-4D as the association strategy.
CA-Net is trained following the official implementation and its recommended training setup.
As shown in Table~\ref{tab:detect_and_track}, Geo-4D consistently improves association quality under identical frame-wise predictions, demonstrating that the gain comes from our association method rather than stronger segmentation masks, consistent with the findings reported in the main paper.

\subsection{Analysis of Global Geometry-aware Soft Matching.}
In the main paper, we argue that the proposed Global Geometry-aware Soft Matching~(GGSM) improves the robustness of the matching strategy against noisy measurements and predictions.
To isolate the contribution of GGSM, we additionally evaluate its effectiveness on non-rigid point cloud registration benchmarks~\citep{li2022lepard, li2022non} independent of tracking performance.
This allows us to analyze whether GGSM improves geometric correspondence estimation under challenging conditions such as partial overlap, occlusion, and large motion.

\myparagraph{Baselines.}
While more advanced baselines exist for non-rigid point cloud registration, we compare against several representative registration methods to better understand the effectiveness of GGSM.
Specifically, we include Iterative Closest Point (ICP)~\citep{besl1992method} as a classical baseline, Sinkhorn-based registration which is closely related to our global geometry association, Non-rigid ICP~(NICP)~\citep{newcombe2015dynamicfusion} as an extension of ICP for non-rigid registration, and Neural Scene Flow Prior (NSFP)~\citep{li2021neural} which predicts point-wise motion using neural networks.
ICP determines the global rotation and translation parameters, which are designed for rigid transformations.
Optimal Transport~(OT)~\citep{cuturi2013sinkhorn} iteratively estimates point displacements by optimizing a transport plan via the Sinkhorn-Knopp algorithm~\citep{sinkhorn1967diagonal} and updating point locations according to the gradients of the OT objective.
NICP estimates the movement of all points individually rather than relying on a single rigid transformation. We adopt a typical solution that optimizes a deformation graph with regularization constraints to enforce smooth deformations.
NSFP optimizes a multilayer perceptron~(MLP) to predict point-wise motion by minimizing the Chamfer distance between two point clouds.

\myparagraph{Metrics.}
We adopt widely used metrics to evaluate how well the registered points align with the ground truth. End-Point Error (EPE) measures the average 3D warping error across all points. AccS and AccR represent the percentage of points whose relative errors are below 2.5$\%$ and 5.0$\%$, respectively. Outlier denotes the percentage of points whose relative error exceeds 30$\%$. 
EPE and Outlier evaluate the overall registration performance, while AccS and AccR measure the point-wise accuracy.

\myparagraph{Experimental Analysis.}
As shown in Tab.~\ref{tab:4dmatch_joint}, two key observations emerge across both 4DMatch-F and 4DLoMatch-F.
First, GGSM achieves the lowest EPE and Outlier rates despite relying only on rigid transformations.
Although non-rigid methods achieve higher point-wise accuracy (AccS and AccR), they often introduce globally inconsistent deformations, which can degrade the overall geometric alignment.
In contrast, GGSM leverages global geometric structure to estimate a coherent transformation, leading to more reliable overall registration performance.
Second, GGSM provides a well-balanced trade-off between efficiency and performance.
NSFP models point-wise motion by optimizing a neural network for each point coordinate, which offers strong flexibility but incurs significant computational overhead.
Moreover, NICP estimates per-vertex deformation parameters through iterative non-rigid optimization, which also limits efficiency.
On the other hand, our GGSM remains computationally efficient as it inherits the efficiency of rigid ICP, resulting in strong overall registration quality by improving geometric consistency.

\begin{table}[t!]
\centering
\small
\caption{Quantitative comparison on 4DMatch-F~\citep{li2022non} and 4DLoMatch-F~\citep{li2022non}.}
\setlength{\tabcolsep}{5pt}
\resizebox{\textwidth}{!}{
\begin{tabular}{lcccc|ccccc}
\toprule
& \multicolumn{4}{c}{\textbf{4DMatch-F}} 
& \multicolumn{4}{c}{\textbf{4DLoMatch-F}} \\
\cmidrule(lr){2-5} \cmidrule(lr){6-9}
Method 
& EPE$\downarrow$ & AccS$\uparrow$ & AccR$\uparrow$ & Outlier$\downarrow$
& EPE$\downarrow$ & AccS$\uparrow$ & AccR$\uparrow$ & Outlier$\downarrow$ & Time$^*$ \\
\midrule
ICP~\citep{besl1992method}
& 0.296 & 2.96 & 12.06 & 71.50
& 0.565 & 0.14 & 0.74 & 90.87 & 0.10 \\
Sinkhorn~\citep{sinkhorn1967diagonal}
& 0.308 & 2.76 & 8.13 & 79.86
& 0.505 & 0.20 & 0.86 & 89.47 & 3.76 \\
NICP~\citep{newcombe2015dynamicfusion}
& 0.325 & 6.44 & 12.10 & 80.40 
& 0.517 & 0.18 & 0.73 & 92.37 & 4.80 \\
NSFP~\citep{li2021neural}
& 0.265 & \textbf{8.66} & \textbf{18.65} & 64.96 
& 0.495 & \textbf{0.38} & \textbf{1.56} & 84.77 & 39.54 \\
\rowcolor{gray!20}\textbf{Ours (ICP+GGSM)}
& \textbf{0.210} & 2.51 & 11.26 & \textbf{62.58} 
& \textbf{0.431} & 0.18 & 1.19 & \textbf{78.80} & 1.62 \\
\bottomrule
\end{tabular}
}
\label{tab:4dmatch_joint}
{\tiny \hfill $^*$ Runtime measured on a single NVIDIA A100.}
\end{table}

\myparagraph{Takeaway:}
In this section, we validate the effect of GGSM in a controlled setting by excluding other factors (e.g., rigid objects and driving surface in on-road environments). 
It is important to note that the primary scope of this work is 4D LiDAR panoptic segmentation in driving scenes, rather than non-rigid point cloud registration itself. 
These experimental results support our claim that GGSM plays a crucial role in enabling robust and reliable instance association in driving scenes with both rigid and non-rigid instances. For additional discussions, please refer to the corresponding analyses presented in the main manuscript.

\subsection{Analysis of DBSCAN Instance Refinement.}
In this specific experiment of integrating \textsc{Geo-4D} with Mask4Former, we apply DBSCAN clustering~\citep{campello2013density} as done in the original Mask4Former setting.
Table~\ref{tab:dbscan_ablation} summarizes the results, where we derive two major insights.
First, \textsc{Geo-4D} achieves performance comparable to existing state-of-the-art methods without the additional clustering step, and further surpasses them when clustering is applied.
Furthermore, we discover that the increase of runtime primarily results from the post-processing step.
\begin{wraptable}{r}{0.50\textwidth}
\vspace{-5pt}
\centering
\caption{\textbf{Instance refinement.}}
\label{tab:dbscan_ablation}

\scriptsize
\setlength{\tabcolsep}{4.5pt}
\renewcommand{\arraystretch}{1.07}

\begin{tabular}{lcccc}
\toprule
Setting
& LSTQ$\uparrow$
& $S_{\rm assoc}\uparrow$
& $S_{\rm cls}\uparrow$
& Time (s)$\downarrow$ \\
\midrule

w/o DBSCAN
& 71.8
& 75.8
& 67.5
& \textbf{0.34} \\

w/ DBSCAN
& \textbf{72.7}
& \textbf{78.3}
& 67.5
& 1.10 \\

\bottomrule
\end{tabular}
\vspace{-8pt}
\end{wraptable}
Without this step, \textsc{Geo-4D} delivers the most efficient runtime of 340 ms compared to all other methods, with only a marginal loss in performance.
These observations clearly highlight that \textsc{Geo-4D} offers both high efficiency and strong performance, and can further benefit from improvements in the backbone network.

\subsection{Quantitative Results}
\label{app:C4}
We additionally provide the results of each semantic class on panoptic nuScenes~\citep{fong2022panoptic} to facilitate a more comprehensive understanding.
In Table~\ref{tab:app_nusc}, our method yields overall performance improvements, particularly in safety-critical \textit{thing} classes~(e.g., bicycle, car, and pedestrian).
Furthermore, classes with relatively fewer points, such as pedestrians and traffic cones, exhibit stronger association performance against the baselines.
This observation aligns with our earlier analysis on SemanticKITTI~\citep{behley2019semantickitti} presented in the main manuscript.

\begin{table*}[t!]
\scriptsize
\setlength{\tabcolsep}{0.0050\linewidth}

\newcommand{\classfreq}[1]{{~\tiny(\nuscenesfreq{#1}\%)}}

\centering
\caption{Comparison of the quantitative results on panoptic nuScenes~\citep{fong2022panoptic}.}

\resizebox{\textwidth}{!}{
\begin{tabular}{c|ccc|*{10}{c}|*{6}{c}}
\toprule
& & & &
\multicolumn{10}{c|}{\textbf{Thing classes (AQ / IoU)}} &
\multicolumn{6}{c}{\textbf{Stuff classes (IoU only)}} \\
\cmidrule(lr){5-14} \cmidrule(lr){15-20}
Model & LSTQ & S$_{assoc}$ & S$_{cls}$ &
\rotatebox{90}{\textcolor{nbarrier}{$\blacksquare$} barrier} &
\rotatebox{90}{\textcolor{nbicycle}{$\blacksquare$} bicycle} &
\rotatebox{90}{\textcolor{nbus}{$\blacksquare$} bus} &
\rotatebox{90}{\textcolor{ncar}{$\blacksquare$} car} &
\rotatebox{90}{\textcolor{nconstruct}{$\blacksquare$} const. veh.} &
\rotatebox{90}{\textcolor{nmotor}{$\blacksquare$} motorcycle} &
\rotatebox{90}{\textcolor{npedestrian}{$\blacksquare$} pedestrian} &
\rotatebox{90}{\textcolor{ntraffic}{$\blacksquare$} traffic cone} &
\rotatebox{90}{\textcolor{ntrailer}{$\blacksquare$} trailer} &
\rotatebox{90}{\textcolor{ntruck}{$\blacksquare$} truck} &
\rotatebox{90}{\textcolor{ndriveable}{$\blacksquare$} drive. surf.} &
\rotatebox{90}{\textcolor{nother}{$\blacksquare$} other flat} &
\rotatebox{90}{\textcolor{nsidewalk}{$\blacksquare$} sidewalk} &
\rotatebox{90}{\textcolor{nterrain}{$\blacksquare$} terrain} &
\rotatebox{90}{\textcolor{nmanmade}{$\blacksquare$} manmade} &
\rotatebox{90}{\textcolor{nvegetation}{$\blacksquare$} vegetation} \\
\midrule

Mask4Former~\citep{yilmaz2024mask4former} &
67.6 & 63.0 & 72.5 & 48.3/76.9 & 42.6/37.0 & 60.5/90.2 & 76.7/92.7 & 42.7/44.9 & 61.1/77.0 & 53.5/74.0 & 48.9/55.7 & 38.7/47.8 & 62.2/77.8 & 95.6 & 69.2 & 71.5 & 74.6 & 88.5 & 87.0 \\

Mask4D~\citep{marcuzzi2023mask4d} &
58.5 & 54.1 & 63.1 & 29.7/64.3 & 27.8/20.5 & 60.5/79.2 & 72.5/83.5 & 29.5/21.8 & 52.5/72.0 & 42.3/65.9 & 39.5/43.0 & 19.2/35.2 & 58.9/71.8 & 94.2 & 53.9 & 64.9 & 71.3 & 85.2 & 84.5 \\

CA-Net~\citep{marcuzzi2022contrastive} &
63.4 & 56.6 & 71.0 & 30.1/74.5 & 31.9/29.1 & 56.5/88.5 & 73.3/91.2 & 38.6/43.2 & 63.2/78.7 & 53.6/71.2 & 34.6/54.6 & 39.5/48.3 & 62.1/77.3 & 95.6 & 68.4 & 70.3 & 72.7 & 86.7 & 85.2 \\

\rowcolor{gray!20}Geo-4D (Ours) &
\textbf{67.9} & \textbf{65.0} & 71.0 &
40.7/74.5 & 44.1/29.1 & 58.9/88.5 & 80.9/91.2 & 40.1/43.2 &
65.9/78.7 & 67.2/71.2 & 52.0/54.6 & 40.6/48.3 & 65.7/77.3 & 95.6 & 68.4 & 70.3 & 72.7 & 86.7 & 85.2\\
\bottomrule
\end{tabular}
}
\label{tab:app_nusc}
\end{table*}

\subsection{Qualitative Results}
\label{app:C5}
\begin{figure*}[t!]
\begin{center}
   \includegraphics[width=\textwidth]{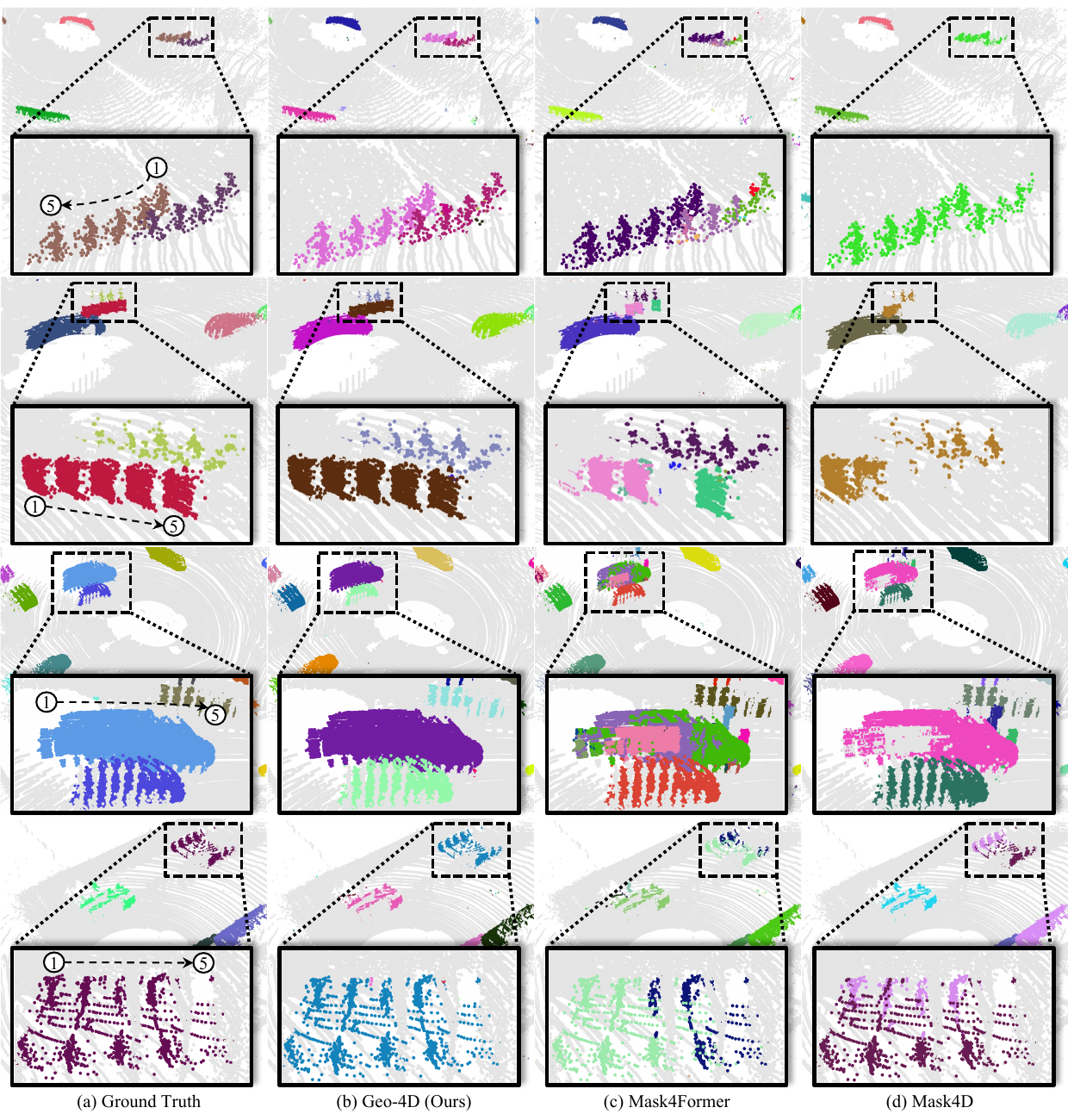}
\end{center}
\caption{Qualitative comparison on semanticKITTI~\citep{behley2019semantickitti} validation set. We visualize the association results over five consecutive scans for both the baselines and our \textsc{Geo-4D}. Different colors represent different instances. The dotted boxes zoom in for clear comparison. For clarity, we use circled numbers (\textcircled{1}–\textcircled{5}) to represent the frame indices.}
\label{fig:fig_qual_add}
\end{figure*}

\begin{figure*}[t!]
\begin{center}
   \includegraphics[width=\textwidth]{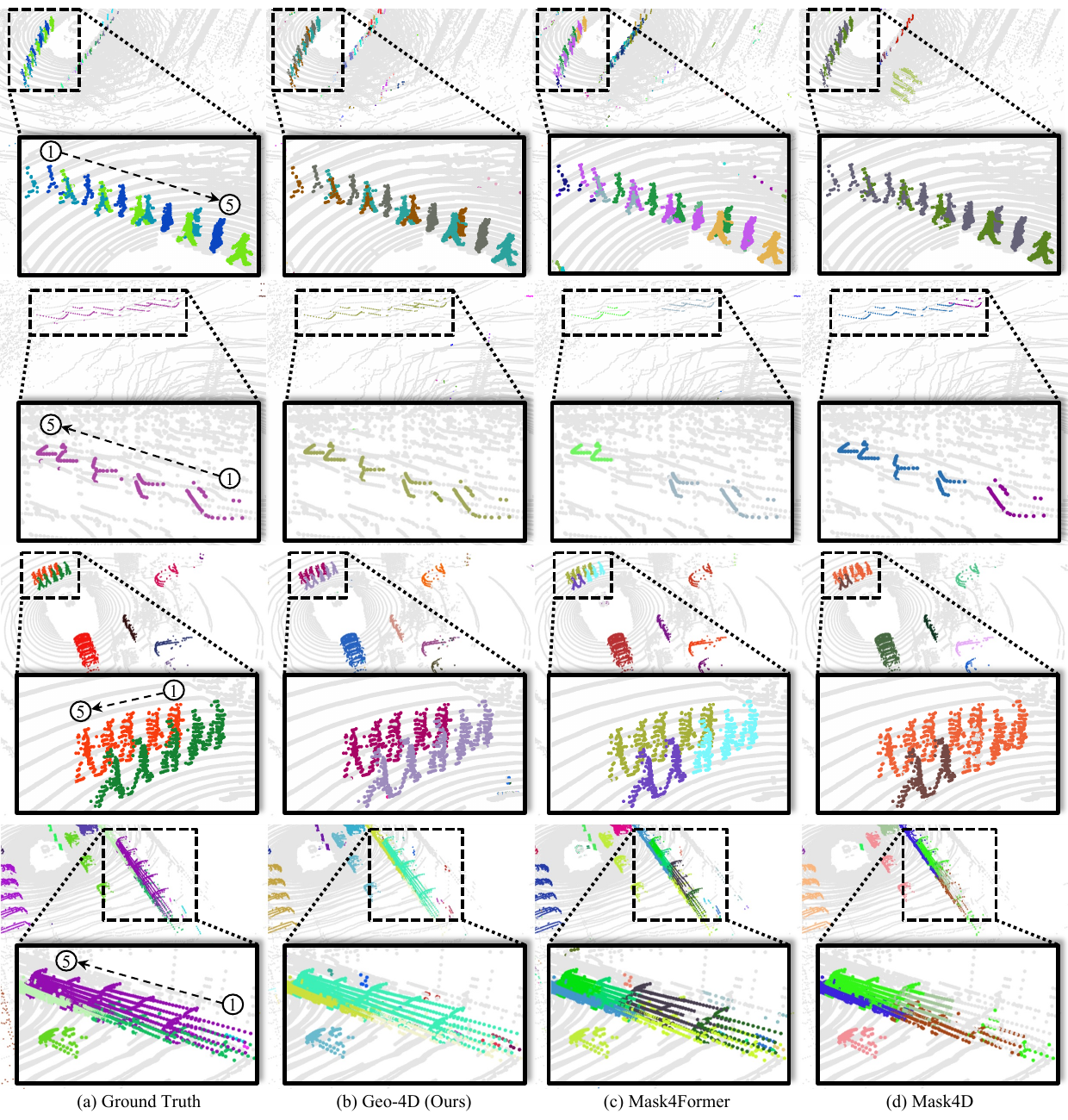}
\end{center}
\caption{Qualitative comparison on panoptic nuScenes~\citep{fong2022panoptic}. We visualize the association results over five consecutive scans for both the baselines and our \textsc{Geo-4D}. Different colors represent different instances. The dotted boxes zoom in for clear comparison. For clarity, we use circled numbers (\textcircled{1}–\textcircled{5}) to represent the frame indices.}
\label{fig:fig_qual_add_nu}
\end{figure*}

In this section, we present additional qualitative analysis by illustrating the results of SemanticKITTI~\citep{behley2019semantickitti} and panoptic nuScenes~\citep{fong2022panoptic} dataset.
As clearly shown in Figure~\ref{fig:fig_qual_add} and ~\ref{fig:fig_qual_add_nu}, \textsc{Geo-4D} maintains robust ID consistency even in challenging scenarios.
Our method explicitly incorporates the inherent geometric relations between instances and adopts a delicately designed pipeline that is resilient to occlusion and noisy predictions.
In contrast, Mask4Former~\citep{yilmaz2024mask4former} associates instances by selecting the majority ID among spatially overlapped points, making it vulnerable to imprecise predictions and leading to frequent ID switches~(see the 1$^{st}$, 2$^{nd}$, and 3$^{rd}$ rows of Figure~\ref{fig:fig_qual_add}). 
Apart from this, Mask4D~\citep{marcuzzi2023mask4d} struggles with separating nearby objects and accumulates errors from early mispredictions~(see the 1$^{st}$, 2$^{nd}$, and 4$^{th}$ rows of Figure~\ref{fig:fig_qual_add}).
Furthermore, \textsc{Geo-4D} also operates well on the panoptic nuScenes dataset, which is sparse and large temporal gaps between consecutive scans.
Specifically, in the 2$^{nd}$ and 4$^{th}$ rows of Figure~\ref{fig:fig_qual_add_nu}, we observe that the instances have no overlap and sparse representations.
The baselines confuse the instance IDs due to their partial observations and rapid motions.
However, \textsc{Geo-4D} successfully finds the corresponding instances even when they are distant.
Also, from the 1$^{st}$ row of Figure~\ref{fig:fig_qual_add_nu}, we observe that \textsc{Geo-4D} can effectively associate instances in complex scene where pedestrians are heavily entangled.

\section{Discussion}
\label{app:D}

\subsection{Limitation and Future work}
\label{app:D1}
We identify several limitations of our proposed approach, which inherits issues from both pre-trained 3D panoptic models and LiDAR sensors.
While \textsc{Geo-4D} demonstrates strong association performance with a cost efficient pipeline, real-time deployment remains challenging.
Our method highly depends on the performance of the underlying 3D panoptic model; thus, its runtime can be further improved as point cloud processing continues to advance.
In addition, partial observation of instances in LiDAR sensing make it difficult to reliably associate instances.
Incorporating trajectory cues represents a promising direction to further stabilize instance association.
Lastly, our pipeline forces the requirement of camera parameters as a prerequisite for transforming the point coordinates into the world coordinates.
Relaxing this requirement would improve the applicability of our method in broader real-world settings.
We hope these insights encourage continued progress toward scalable 4D perception systems.

